\documentclass[mathematics,article,submit,moreauthors]{Definitions/mdpi} 

\def\today{August 8, 2022}
\renewcommand{\linenumbers}{}

\firstpage{1} 
\pubvolume{1}
\issuenum{1}
\articlenumber{0}
\pubyear{2022}
\copyrightyear{2022}
\datereceived{} 
\dateaccepted{} 
\datepublished{} 
\hreflink{https://doi.org/} 

\usepackage{graphicx}
\usepackage[final]{changes}

\Title{Improving Large-Scale $k$-Nearest Neighbor Text Categorization with Label Autoencoders}

\TitleCitation{Improving Large-Scale $k$-nearest neighbor text categorization with label autoencoders}

\Author{Francisco J. Ribadas-Pena 
* 
$^{,\dagger,\ddagger}$\orcidA{}, Shuyuan Cao $^{\ddagger}$ and V\'{\i}ctor M. Darriba Bilbao 
 $^{\ddagger}$}

\AuthorNames{Francisco J. Ribadas-Pena, Shuyuan Cao and V\'{\i}ctor M. Darriba Bilbao}

\AuthorCitation{Ribadas-Pena, F.J.; Cao, S.; Darriba, V.M.}

\address[1] {%
{Department of Computer Science}
, University of Vigo, {36310 Vigo, Spain}
; shuyuan.cao@uvigo.es (S.C.);   darriba@uvigo.es (V.M.D.B.)}

\corres{\hangafter=1 \hangindent=1.05em \hspace{-0.82em}Correspondence: ribadas@uvigo.es}

\firstnote{\hangafter=1 \hangindent=1.05em \hspace{-0.82em}Current address: Edificio Polit\'ecnico, Campus As Lagoas s/n, 32004 Ourense, Spain} 
\secondnote{\hangafter=1 \hangindent=1.05em \hspace{-0.82em}These authors contributed equally to this work.}

\abstract{%
In this paper, we introduce a multi-label lazy learning approach to deal with automatic semantic indexing in large document collections in the presence of complex and structured label vocabularies with high inter-label correlation. 
The proposed method is an evolution of the traditional $k$-Nearest Neighbors algorithm which uses a large autoencoder trained to map the large label space to a reduced size latent space and to regenerate the predicted labels from this latent space.
We have evaluated our proposal in a large portion of the MEDLINE biomedical document collection which uses the Medical Subject Headings (MeSH) thesaurus as a controlled vocabulary. 
In our experiments we propose and evaluate several document representation approaches and different label autoencoder configurations.	
}



\newcommand{\LAE}{\emph{label-AE}}
\newcommand{\LAEs}{\emph{label-AEs}}

\begin{document}
\section{Introduction}

In Large-Scale Text Categorization (LSTC) we are confronted with textual classification problems where a very large and structured set of possible classes are employed. 
For the general case, not limited exclusively to text, the~term eXtreme Multi-Label categorization (XML) is also often used.
Usually, in these cases we are dealing with multi-label learning problems where models learn to predict more than one class or label to be assigned to a given input~text.

Conventional approaches in multi-label learning either convert the original multi-label problem into a set of single-label problems or adapt well known single-label classification algorithms to handle multi-label datasets.
In the context of LSTC and XML research, evolutions of both types of method, that employ what has been called label embedding (LE) or label compression (LC),
have recently emerged, trying to take advantage of label dependencies to improve categorization performance. 
LE methods try to take advantage of label dependencies to improve categorization performance.
The starting premise of LE is to convert the large label spaces to a reduced-dimensional representation space (the embedding space) where the actual classification is performed, the~results of which are then transformed back to the original label~space.

Autoencoders (AEs) are a classical unsupervised neural network architecture able to learn compressed feature representations from original features.
Usually AEs are symmetrical networks with a series of layers that learn to transform their input to a latent space of lower dimension (encoder) and another series of layers that learn to regenerate that input from the latent space (decoder), both of them are connected by a small layer that acts as an information bottleneck.
Training is carried out in an unsupervised way, presenting the same training vector in the input layer and in the output layer. 
%
AE are typically employed in data pre-processing, discarding the decoder and using the learned encoder to create rich representations of the input data useful in further processing.

Automatic semantic indexing is often modeled as an LSTC or XML problem. This task seeks to automate assigning  to a given input document a sets of descriptors or indexing terms taken from a controlled vocabulary in order to improve further searching tasks.
MeSH (Medical Subject Headings) is a large semantic thesaurus commonly used in the management of biomedical literature. MeSH labels are semantic descriptors arranged into 16 overlapping concept sub-hierarchies, which are employed to index MEDLINE, a~collection of millions of biomedical~abstracts.

Given this context, in~this paper, a~multi-label lazy learning approach  is presented to deal with automatic semantic indexing in large document collections in the presence of complex and structured label vocabularies with high inter-label correlation. 
This method is an evolution from the traditional $k$-Nearest Neighbors ($k$-NN) algorithm which exploits an AE trained to map the large label space to a reduced size latent space and to regenerate the output labels from this latent space.
Our contributions are as~follows:
\begin{itemize}
	\item We have employed MEDLINE as a huge labeled collection to train large \LAEs~able to map MeSH descriptors onto a semantic latent space where label interdependence is~coded.
	
	\item Our proposal adapts classical $k$-NN categorization to work in the semantic latent space learned by these AEs and employs the decoder subnet to predict the final candidate labels, instead of applying simple voting schemes like traditional $k$-NN.
	
	\item Additionally, we have evaluated different document representation approaches, using both sparse textual features and dense contextual representations. We have studied their effect in the computation of inter-document distances that are the starting point to find the set of 
	neighbor documents employed in  $k$-NN classification.  
\end{itemize}

The remainder of this article is organized as follows. Section~\ref{sec:related_work} presents the background and context of this paper. Section~\ref{sec:methods} describes the details of the proposed method and its components. Finally, Section~\ref{sec:results} discusses the experimental results obtained by our proposals and Section~\ref{sec:conclusions} summarizes the main conclusions of this~work.

\section{Related~Work}~\label{sec:related_work}
This work is framed at the confluence of three research fields: (1) large-scale multi-label categorization, (2) autoencoders and (3) semantic indexing. This section provides a brief review of the most relevant contributions in the state of the art of these topics in relation to our label autoencoder~proposal.

\subsection{Multi-Label Categorization}
In multi-label learning~\cite{tsoumakas2007multi} examples can be assigned simultaneously to several not mutually exclusive classes. 
This task differs from single-label learning (binary or multi-class) and has its own characteristics that make it more complex, while being able to model many relevant real-world problems.
Formally, given $L = \{l_1, l_2, \dots, l_l\}$ the finite set of labels in a multi-label learning task and $D = \{(x_1, y_1), (x_2, y_2), \dots, (x_n, y_n)\}$ the set of multi-label training instances, where $x_i$ is the $i$-example feature vector and $y_i \subseteq L$ is the set of labels for that example, the~ multi-label categorization task aims to build a multi-label predictor $f : x' \longmapsto y'$, with~$y' \subseteq L$, able to produce good classifications on incoming test instances from $T=\{x'_1, x'_2, \dots, x'_m\}$.

The scientific literature on multiple-label learning~\cite{madjarov2012extensive, zhang2013review} usually identifies two main groups of approaches when dealing with this problem: algorithm adaptation methods and problem transformation methods.
Algorithm adaptation approaches extend and customize single-label machine learning algorithms in order to handle multi-label data directly.
Several adaptations of traditional learning algorithms have been proposed in the literature,
such as boosting (AdaBoost.MH)~\cite{schapire2000boostexter}, 
support vector machines (RankSVM)~\cite{elisseeff2001kernel},
multi-label $k$-nearest neighbors (ML-kNN)~\cite{zhang2007ml} and
neural networks~\cite{zhang2006multilabel}.
On the other hand, problem transformation methods transform a multi-label learning problem into a series of single-label
problems which already have well-established resolution methods. The~solutions of these problems are then combined to solve the original multi-label learning task.
For example, Binary Relevance (BR)~\cite{boutell2004learning}, Label Powerset (LP)~\cite{tsoumakas2010random} and Classifier Chains (CC)~\cite{read2011classifier} transform multi-label learning problems into binary classification~problems.

A relevant aspect in multi-label learning approaches is the treatment given to inter-label dependencies.
The simplest methods, such as BR, do not take into account correlation between labels, assuming label independence and neglecting the fact that some labels are more likely to co-exist. This assumption brings advantages in parallelization and training efficiency, but~at the cost of lower performance in many real-word tasks that exhibit complex inter-label dependencies. 
Other approaches, like LP and CC, try to capture the dependencies between labels using different strategies. For~example, CC sequentially creates a set of binary classifiers where labels predicted by previous classifiers are part of the features employed in successive~classifications.

Recent research in multi-label learning propose a label embedding (LE) or label compression (LC) approach that tries to properly exploit correlation between label information by transforming the label space into a latent label space of reduced dimensionality. The~actual categorization is performed in that latent space, where correlation between labels is implicitly exploited and a proper decoding process will map the projected data back onto the original label space.
Early work in LE~\cite{hsu2009multi,tai2012multilabel} typically considered linear embedding functions and worked with fairly small label space sizes. Other approaches overcome the limitations of linear assumptions evolving to non-linear embeddings~\cite{cisse2013robust,bhatia2015sparse,rai2015large}, including several methods based on conventional or deep neural networks~\cite{wicker2016nonlinear,yeh2017learning, wang-etal-2019-ranking}.

Finally, a~prominent field in multi-label learning that have been attracting lots of research in recent times is eXtreme Multi-label Classification (XML)~\cite{agrawal2013multi,prabhu2014fastxml}. XML is a multi-label classification task in which learned models automatically label a data sample with the most relevant subset of labels from a large label set, with~sizes ranging from thousands to millions. 
This is a challenging problem due to the scale of the classification task, label sparsity and complex label correlations. 
The ability to handle label correlations and the scalability of LE approaches~\cite{bhatia2015sparse} have shown many advantages in XML making embeddings one of the most popular approaches for tackling XML~problems.

\subsection{Autoencoders in Multi-Label Learning}

\begin{figure}[H]
	\begin{center}
\hspace{-80pt} {}\includegraphics[width=0.8\linewidth]{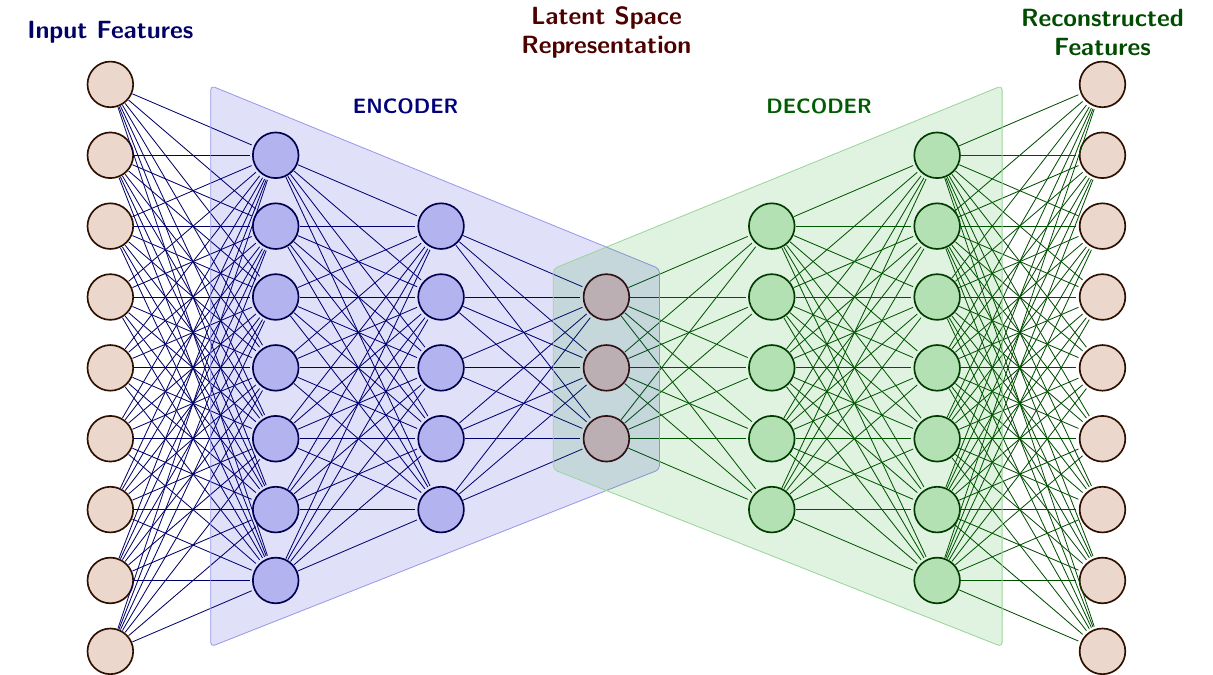}
	\end{center}
	\caption{\label{fig:autoencoder} Architecture of a generic~autoencoder.}	
\end{figure}

Autoencoders (AEs)~\cite{liu2017survey,charte2018practical} are a family of unsupervised 
feedforward Neural Network architectures that jointly learn an encoding function, 
which maps an input to a latent space representation, and~a decoding function, which 
maps from the latent space back onto the original space. Figure~\ref{fig:autoencoder} shows this
symmetric encoder-decoder structure, with;
\begin{itemize}
	\item An encoder function $Enc: X \to Z$, which maps the input vectors into a latent (often lower-dimensional) representation  though a set of hidden layers.
	\item A decoder function $Dec: Z \to X$, which acts as an interpreter of the latent representation and reconstructs the input vectors though a set of hidden layers, usually symmetric with the encoding layers.
	\item A middle hidden layer representing in the latent space $Z$ an encoding of the \mbox{input data.} 
\end{itemize}

Training the model to reproduce the input data at its output, AE jointly optimizes the parameters of encoder $Enc$ and decoder $Dec$ functions and 
can learn in its hidden layer richer non-linear encoding features that can 
represent complex input data in a reduced dimensionality latent space.
Most practical applications of AE exploit this latent representation 
in (1) data compression or hashing tasks, (2) classification tasks, using AE to reduce input features dimensionality with minimal information loss,
(3) anomaly detection, by~analyzing outliers and abnormal patterns in generated embeddings, (4) visualization tasks on the encoded space or (5) data reconstruction and noise reduction in image~processing.

Usage of AEs in single-label and multi-label learning, including XML, has already been reported in many research works and AEs
are frequently part of pre-processing steps performing dimensionality reduction
in order to improve categorization performance and speed. Methods like AE-$k$NN~\cite{aeknn} train an AE on high dimensional input features from a training dataset and employ the encoder sub-net as an input feature compressor, transforming the original input space into a lower-dimensional one where a conventional instance-based $k$-NN algorithm does the~labeling.

The use of AEs over the label space  has been less common in the literature.
With the advent and explosion of XML methods, research proposals that try to take advantage of the capabilities of AEs
to capture non-linear dependencies among the labels have~appeared.

Wicker et al.~\cite{wicker2016nonlinear}, in~a pioneering work in the use of label space AEs,
introduce MANIC, a~multi-label classification algorithm following the problem transformation
approach which extracts non-linear dependencies between labels by compressing them using an AE. 
MANIC uses the encoder part to replace training labels with a reduced dimension version and 
then trains a base classifier (a BR model in their proposal) using the compressed labels as new target~variables.

C2AE (Canonical-Correlated AutoEncoder)~\cite{yeh2017learning} and Rank-AE (Ranking-based Auto-Encoder)~\cite{wang-etal-2019-ranking} follow a similar idea, which was later generalized in~\cite{Jarrett2020Target-Embedding}.
These approaches perform a joint embedding learning by deriving a compressed feature space shared by input features and labels.
An input space encoder and a label space encoder, sharing the same hidden space, and~a decoder
that converts this hidden space back to the original label space are trained together to create 
a deep latent space that embeds input features and labels simultaneously.
%



\subsection{Semantic Indexing in the Biomedical Domain}

Controlled vocabularies provide an efficient way of accessing and organizing large collections of textual documents, specially in domains where a simple text-based representation of information is too ambiguous, like the biomedical or legal domains. Automatic semantic indexing seeks to build  systems able to annotate an arbitrary piece of text with relevant controlled vocabulary terms. 
Aside from pure natural language processing (NLP) based methods, most of them following Named Entity Recognition (NER) or Entity Linking strategies, many approaches to semantic indexing model the assignment of controlled vocabulary terms as a multi-label categorization~problem.

The Medical Subject Headings (MeSH) thesaurus~\cite{mesh} is a controlled and hierarchically-organized vocabulary, developed and maintained by the National Library of Medicine ({\url{https://www.nlm.nih.gov/}}) 
, which was created for categorizing and searching citations in MEDLINE and the PubMED database. 
{MEDLINE} 
 (\url{https://www.nlm.nih.gov/medline/medline_overview.html}) 
 is an NLM bibliographic database that contains more than 29 million references to journal articles in life sciences published from 1966 to present, published in more than 5200 journals worldwide in about 40 languages. Each MEDLINE citation contains the title and abstract of the original article, author information, several metadata items (journal name, publishing dates, etc.) and a set of MeSH descriptors that describe the content of the citation and were assigned by NLM annotators to help indexing and searching MEDLINE records.
The task of identifying the MeSH terms that best represent a MEDLINE article is manually performed by human experts. 
The average number of descriptors per citation in MEDLINE 2021 edition was~12.68.

MeSH vocabulary is composed of MeSH subject headings (commonly known as descriptors) which describe the subject of each article and a set of standard qualifiers (subheadings) which narrow down the MeSH heading topic.
Additionally, Check Tags are a special subset of 32 MeSH descriptors that cover concepts mentioned in almost every article (human age groups, sex, research animals, etc.). 
MeSH descriptors are arranged in a hierarchy with 16 top-level categories that constitute a collection of overlapping topic sub-thesauri. A~given descriptor may appear at several locations in the hierarchical tree and can have several broader terms and several narrower terms.
The 2021 edition of the MeSH thesaurus is composed of 29,917 unique headings, hierarchically arranged in one or more of the 16 top-level categories, with~the distribution shown in Table~\ref{table:mesh}.

	\begin{table}[H]
	\caption{{Descriptor distribution} 
 in MeSH top-level categories.}
\newcolumntype{C}{>{\centering\arraybackslash}X}
\begin{tabularx}{\textwidth}{m{6cm}<{\raggedright}b{7cm}<{\raggedleft}}
\toprule
				& \textbf{\# of }\\ 
				\textbf{Subhierarchy} & \textbf{Descriptors}\\ ine
				(A) Anatomy  & 2927\\
				(B) Organisms& 5196\\
				(C) Diseases & 11,303\\
				(D) Chemicals and Drugs & 20,992\\
				(E) Analytical, Diagnostic and  & 4764\\
				\hspace*{12pt} Therapeutic Techniques   &    \\
				\hspace*{12pt} and Equipment &    \\
				(F) Psychiatry and Psychology & 1150\\
				(G)  Biological Sciences & 3428\\
				(H) Physical Sciences & 513\\

				(I) Anthropology, Education,  & 651\\
				\hspace*{12pt} Sociology and  & \\ 
				\hspace*{12pt} Social Phenomena & \\ 
				(J) Technology and Food  & 601\\
				\hspace*{12pt}     and Beverages & \\
				(K) Humanities & 218\\
				(L) Information Science & 519\\
				(M) Persons & 258\\
				(N) Health Care & 2350\\
				(V) Publication Characteristics & 188\\
				(Z) Geographic Locations & 553\\
			\bottomrule
\label{table:mesh}
		\end{tabularx}
	\end{table}

\vspace{-8pt} {}
Automatic indexing with MeSH poses great research challenges.  
(1) Beyond its large size, the~distribution of descriptors follows a power-law, where a few labels (Check Tags and very general descriptors) appear in a large number of citations, whereas most descriptors are employed to annotate very few abstracts~{According to statistics} in~\cite{dai2020fullmesh}, only 1\% of all \linebreak{}MeSH headings which have more than 5000 occurrences contribute to more than 40\% of indexing.
(2) Simultaneously  indexing within 16 top-level overlapping topic sub-hierarchies leads to complex label~interdependency. 

Over the years several proposals have attempted to tackle the problem of automatic MeSH indexing.
The Medical Text Indexer (MTI)~\cite{mork201712} is a tool in permanent development by NLM for internal usage as a preliminary 
annotation tool of incoming MEDLINE citations. MTI is based on a combination of NLP based concept finding performed with MetaMap~\cite{aronson2010overview}, $k$-NN prediction using descriptors from PubMed-related citations and several hand-crafted rules and state-of-the-art machine learning techniques that have been incorporated over years of~development.

Semantic indexing with MeSH descriptors has also been boosted in recent years by competitions such as the BioASQ challenge~\cite{tsatsaronis2015overview}, which, since 2013, has been organizing an annual shared-task dedicated to semantic indexing in MEDLINE. 
Several state-of-the-art methods for MeSH indexing were introduced by teams participating in this challenge, most of them modeling the task as
a multi-label learning problem~\cite{gargiulo2019deep}. Some relevant recent developments in MeSH indexing are MeSHLabeler~\cite{liu2015meshlabeler}, DeepMeSH~\cite{peng2016deepmesh}, MeSH Now~\cite{mao2017mesh},  AttentionMeSH~\cite{jin2018attentionmesh}, MeSHProbeNet~\cite{xun2019meshprobenet}, FullMeSH~\cite{dai2020fullmesh}, BERTMeSH~\cite{10.1093/bioinformatics/btaa837} and $k$-NN methods using ElasticSearch and MTI such as~\cite{bedmar2017search}.

\section{Materials and~Methods}~\label{sec:methods}
Our proposal models semantic indexing over MeSH as a multi-label categorization problem with the following specific~characteristics:
\begin{itemize}
	\item MEDLINE provides us with an extensive collection of manually annotated documents to train our models.
	\item MeSH is a rich hierarchical thesaurus with a large set of descriptors and complex label co-occurrence.
\end{itemize}

The approach that we describe in this work tries to take advantage of these characteristics through the use of label autoencoders (\LAEs).
Our method starts by training a \LAE~using the historical information available in MEDLINE.
Once trained, the~components of this AE allow us to (1) convert the MeSH descriptors assigned to a MEDLINE citation to an embedded semantic space using the encoder part and (2) use the decoder part and a simple threshold scheme to return from that reduced-dimensional space back to the MeSH descriptor~space.

The proposed multi-label classification follows a label embedding approach. Our method aims to take advantage of the reduced-dimensional semantic space learned by the \LAE\ so that a lazy learning scheme operates on it performing the actual classification. This results in a $k$-NN classifier  enriched with the compact semantic information provided by the AE components.
This section details the elements that make up our~proposal.

\subsection{Similarity Based Categorization ($k$-NN)}

The $k$-Nearest Neighbor ($k$-NN) algorithm~\cite{aha1991instance} is a lazy learning method which classifies new samples 
using previous classifications of similar samples assuming the new ones will fall into the same or similar categories. 
For a given test instance, $x$, the~$k$ most similar instances (the $k$-nearest neighbors), denoted as $N(x)$, are taken from the training set according to a certain similarity measure. Votes on the labels of instances in $N(x)$ are taken to determine the predicted label for that test instance $x$.

Approaches based on $k$-NN have been widely used in large-scale multi-label categorization in many domains, including MEDLINE documents~\cite{bedmar2017search, trieschnigg2009mesh, ribadas2021cole}. This preference for this lazy learning method is mainly due to its scalability, minimum parameter tuning and,  despite its simplicity, its ability to deliver acceptable results when a large number of training samples are~available.

The basic $k$-NN method we employ in our proposal follows these~steps:
\begin{enumerate}
	\item Create an indexable representation from the textual contents of every document (MEDLINE citations in our case) in the training~dataset.
	
	Two different approaches for creating these indexable representations, dense and sparse, were evaluated in our study as is shown in Section~\ref{sec:representations}.
	
	\item Index these representations in a proper data structure in order to efficiently query it to retrieve sets of similar~documents.
	
	\item For each new document to annotate, the~created index is queried using the indexable representation of the new~document. 
	
	The list of similar documents retrieved in this step together with their corresponding similarity measures are used to determine the following~results:
	\begin{enumerate}[leftmargin=7.mm,labelsep=4.0mm]
		\item expected number of labels to assign to the new document
		\item ranked list of predicted labels (MeSH descriptors in our case)
	\end{enumerate}      
\end{enumerate}

The first aspect conforms to a regression problem, which aims to predict the number of labels to be included in 
the final list, depending on the number of labels assigned to the most similar documents identified during the query phase and on their respective similarity scores.
In our method the number of labels to be assigned is predicted by simply averaging the length of label lists in neighbor~samples.

The other task is a multi-label classification problem, which aims to predict an output label list based 
on the labels manually assigned to the most similar documents.
Our method creates the ranked list of labels using a simple majority voting scheme. 
Since this is actually a multi-label categorization task, there are
as many voting tasks as there were candidate 
labels extracted from the neighboring documents retrieved by the indexing data structure. 
For each candidate label, positive votes come from similar documents annotated with it and negative votes come from neighbors not including it.
The topmost candidate labels are returned as classification~output.

\subsection{Document~Representation}~\label{sec:representations}
As noted in the preceding section, our proposal indexes representations of the training documents in order to implement the similarity function that provides the set of neighbors and their similarity scores.
In this work we have evaluated two different approaches in document representation, which determine their respective indexing and query schemes, together with document~preprocessing.
\begin{itemize}
	\item Sparse representations created by means of several NLP based linguistically motivated index term extraction methods, employed  as discrete index terms in an Apache Lucene index~({\url{https://lucene.apache.org/}).}
	\item Dense representation created by using contextual sentence embeddings based on Deep Learning language models, stored  in a numeric vector index.
\end{itemize}

\subsubsection{Sparse~Representations}
This approach is essentially a large multi-label $k$-NN classifier backed by 
an Apache Lucene index. 
Lucene is an open-source indexing and searching engine, that implements a vector space model for Information Retrieval,
providing several similarity ranking functions, such as BM25~\cite{robertson1995okapi}. 
Textual content of training documents is preprocessed in order to extract a set of discrete index terms which Lucene conveniently stores in an inverted index. 
When labeling, text from new documents is preprocesed and the extracted index term are treated as query terms
and linked together using a global OR operator to conform the final query sent to the indexing engine to retrieve the most similar documents and their corresponding similarity~scores.

In our case, we have employed the BM25 similarity function. The~scores provided by the indexing engine are similarity measures resulting from the engine's internal computations and the 
weighting scheme being employed, which do not have a uniform and predictable upper bound.
In order for these similarity scores to behave like a real distance metric, we have applied a normalization procedure, that transforms them into a pseudo-distance in $[0,1]$.

Regarding sparse document representation we have evaluated several linguistically motivated index term extraction approaches as introduced in~\cite{ribadas2021cole} for a similar problem in Spanish.
We employed the following methods:
\begin{description}
	\item [{Stemming based representation} 
 (STEMS).] This was the simplest approach which employs stop-word removal, using a standard stop-word list 
	and the  default stemmer from the Snowball project ({\url{http://snowball.tartarus.org}}). 
	
	\item [Morphosyntactic based representation (LEMMAS).] In order to deal with morphosyntactic variation
	we have employed a lemmatizer to identify lexical roots and we also replaced stop-word removal 
	with a content-word selection procedure based on part-of-speech (PoS) tags.
	
	{We have delegated} 
 the linguistic processing tasks to the tools provided by the spaCy Natural Language Processing (NLP) toolkit~({\url{https://spacy.io/}}). In~our case we have employed the PoS tagging and lemmatization information provided by spaCy, using the biomedical English models from the {ScispaCy project} 
~(\url{https://allenai.github.io/scispacy/}).
	
	{Only lemmas from} 
 tokens tagged as a noun, verb, adjective, adverb or as unknown words are taken into account to constitute the final document representation, since these PoSs are considered to carry most of the sentence~meaning.
	
	\item [Noum phrases based representation (NPS).] In order to evaluate the contribution of more powerful NLP techniques,
	we have employed a surface parsing approach to identify syntactic motivated nominal phrases from which meaningful multi-word index 
	terms could be~extracted. 
	
	{Noun Phrase (NP) chunks} 
 identified by spaCy are selected and the lemmas of constituent tokens are joined together to create a multi-word index~term.

	\item [Dependencies based representation (DEPS).] We have also employed as index terms triples of dependence-head-modifier extracted by the dependency parser provided by~spaCy.
	
	{In our case the} 
 spaCy dependency parsing model identifies syntactic dependencies following the Universal Dependencies(UD) scheme. The complex index terms were extracted from the following UD relationships~{Detailed list of UD relationships ({available at} 
 \url{https://universaldependencies.org/u/dep/}}): {\emph{acl}} 
, \emph{advcl}, \emph{advmod}, \emph{amod}, \emph{ccomp}, \emph{compound}, \emph{conj}, \emph{csuj}, \emph{dep}, \emph{flat}, \emph{iobj}, \emph{nmod} , \emph{nsubj}, \emph{obj}, \emph{xcomp}, \emph{dobj} and \emph{pobj}.
	
	\item [Named entities representation (NERS).] Another type of multi-word representation taken into account is named entities. We have employed the NER module in spaCy and the ScispaCy models to extract general and biomedical named entities from document~content.
	
	\item [Keywords representation (KEYWORDS).] The last kind of multi-word representation we have included is keywords extracted with statistical methods from the textual content of articles. We have employed the implementation of the TextRank algorithm~\cite{mihalcea2004textrank} provided by the {textacy library} 
~({\url{https://textacy.readthedocs.io}}).

\end{description}

\subsubsection{Dense~Representations}

The recent rise of powerful contextual language models such as BERT and similar approaches have boosted the performance of multiple language processing tasks and Transformer based solutions dominate the state-of-the-art in many NLP areas. A~natural evolution of these contextual word embeddings is to move them towards embeddings at the sentence-level with approaches such as those in the Sentence Transformers~\cite{reimers-2019-sentence-bert} {project} 
~({\url{https://www.sbert.net/}}) that provides pre-trained models to convert sentences in natural languages into fixed-size dense vectors with enriched~semantics.

We have taken advantage of dense semantic representations of whole sentences as a basis for converting a search for similar documents into a search for similar vectors in the dense vector space where documents from the training dataset are~represented.

We have employed the \texttt{sentence-transformers/allenai-specter} {model} 
~({\url{https://huggingface.co/sentence-transformers/allenai-specter}}) to represent a given MEDLINE abstract as a dense vector. This  
is a conversion of the AllenAI SPECTER model~\cite{cohan-etal-2020-specter}, originally trained to estimate the similarity of two publications, to~SentenceTransformers, which exploits the citation graph to generate document-level embeddings of scientific documents. This model returns a 768-dimension vector from inputs in the form \verb!paper[title] + '[SEP]' + paper[abstract]!.

Once we have the dense representations of the training documents using this procedure, we use the FAISS~\cite{johnson2019billion} {library} 
~(\url{https://github.com/facebookresearch/faiss}) to create a searchable index of these dense vectors. This index allows us to efficiently calculate distances between dense vectors and determine for the dense vector associated with a given test document (our query vector) the list of $k$ closest training dense vectors using the Euclidean distance or other similarity~metrics.

With this mechanism of similarity between dense vectors we can apply the $k$-NN classification procedure described previously. In~this case we can use the real distances provided by the FAISS library between the query vector generated from the text to be annotated and the most similar $k$ dense vectors~directly.

\subsection{Label~Autoencoders}~\label{sec:knn_ae}
Our proposal is a special case of eXtream Multi-Label categorization (XML) using a label embedding approach. In~our case a lazy learning method works on a low dimensional projection of the label space build with a label autoencoder (\LAE).
%

Our method is similar to MANIC~\cite{wicker2016nonlinear}. Both of them learn a conventional AE.
In MANIC the encoder is applied to the entirety of labels from the training examples and uses thresholds to convert their embeddings to a smaller binary label space in which Binary Relevance classifiers are trained. In~our case the encoder only acts on the subset of training examples that are part of the neighbors set, $N(x)$. The~embedded vectors of neighbors are averaged and the decoder transforms that average vector to the original label space.
The AEs used by C2AE~\cite{yeh2017learning} and Rank-AE~\cite{wang-etal-2019-ranking} are very different to ours. They jointly train two input subnets that share the inner embedding layer, one generates embeddings from the input features and the other one generates the same embedding space from labels, These two subnets are trained together with an output subnet that decodes the reduced embedding space to the actual label space. In~the annotation phase, only the subnet that creates the embedding from input features and the decoding subnet are~employed.

The first step of our proposal involves training a \LAE\ using the set of labels taken from the training samples. In~the experiments reported in this paper, those labels are the lists of MeSH descriptors assigned to the MEDLINE citations in our training dataset.
For MeSH, this results in a very large \LAE, with~$>$29 K units in the input layer and another $>$29 K output neurons. Also, input and output vectors are extremely sparse, with~an average of 12 values set to 1. On~the other hand, the~set of training samples is very large and can reach several million if the entire MEDLINE collection is~used.

A tentative preliminary study was performed on a portion of the MEDLINE and MeSH datasets. 
In those preliminary runs we assessed the reconstruction capability of the trained AEs. As~a result, the~topology and main parameters of the \LAE\ scheme used in the experiments reported in this paper have been defined as~follows:
\begin{itemize}
	\item	Encoder with 2 hidden layers of decreasing size.
	\item	Inner hidden embedding layer.
	\item	Decoder with 2 hidden layers of increasing size, symmetrical to the encoder.
	\item	ReLU (Rectified Linear Unit) activation function in hidden layer neurons.
	\item	Feed-forward fully connected layers with a 0.2 Dropout in each hidden layer and batch normalization.
	\item	Output layer with SIGMOID activation function (operating as a multi-label classifier).
	\item	Binary cross-entropy loss function.
\end{itemize}

The second step of our method is to extract the internal representations for the training documents and store them in the corresponding index. As~is shown in Section~\ref{sec:representations} an Apache Lucene textual index is employed for NLP based sparse representations and an FAISS index stores the dense contextual vector~representations.

\begin{figure}[H]
	\begin{center}
		\includegraphics[width=0.98\linewidth]{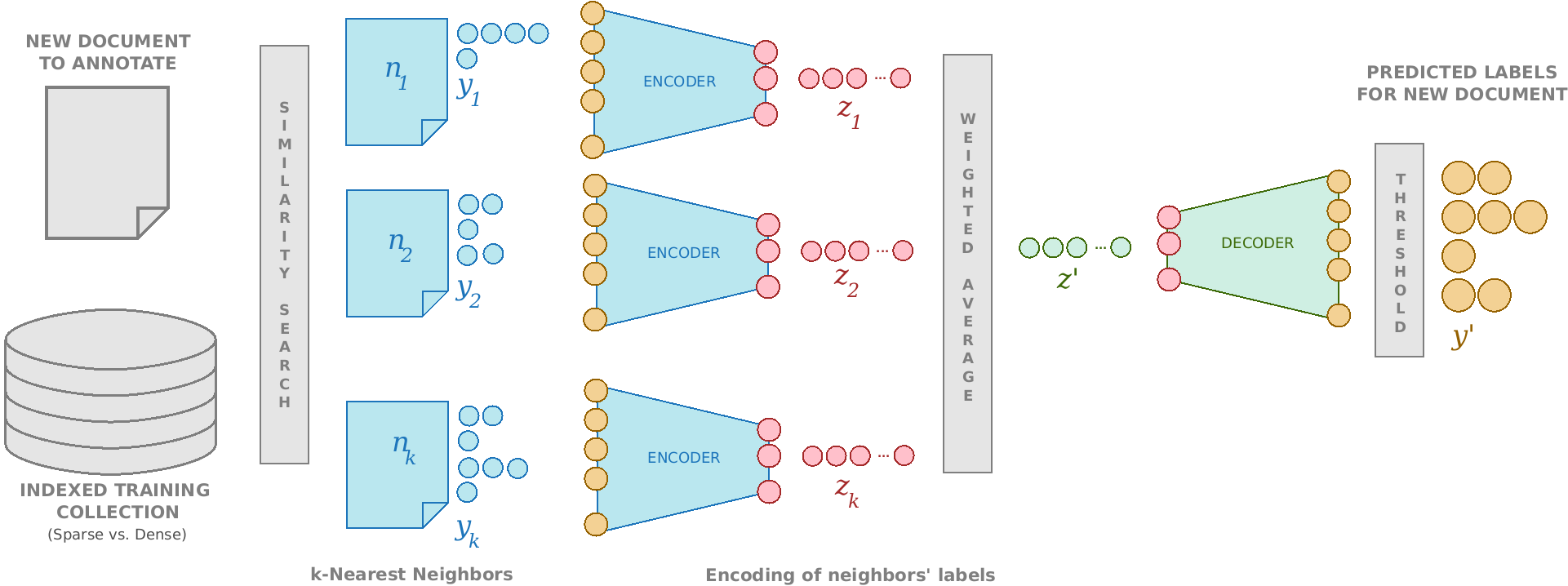}
	\end{center}
	\caption{\label{fig:label_ae_knn} {Categorization} 
 using $k$-NN with label~autoencoders.}	
\end{figure}

Once we have trained our \LAE\ and a properly indexed version of the training dataset is available, to~annotate a new MEDLINE citation $x$, we apply the following procedure, illustrated in Figure~\ref{fig:label_ae_knn}:
\begin{enumerate}
	\item The index is queried and the set $N(x)=\{n_1, n_2, \dots, n_k\}$ with the $k$ documents closest to $x$ is retrieved, along with their respective distances to $x$, ($d_i$ for each $n_i \in N(x)$).
	\begin{itemize}
		\item Depending on the representation being used, title and abstract of $x$ are converted into a sparse set of Lucene indexing terms or into a dense~vector.
		
		\item Once the respective index (Lucene or FAISS) is queried, an~ordered list of most similar citations is available, together with an estimate of their distances to the query document $x$.
		\begin{itemize}[leftmargin=7.mm,labelsep=4.0mm]
			\item BM25 scores converted to a pseudo-distance in $[0,1]$ with Lucene index
			\item euclidean distance between dense representations with FAISS index
		\end{itemize}
	\end{itemize}
	\item The encoder is applied to translate the set of labels assigned to each neighbor $n_i \in N(x)$ into the reduced semantic space, computing $\vec{z_i} = Enc(y_{n_i}))$ $\forall x_i \in N(x)$, with~$y_{n_i}$ the set of labels in neighbor $n_i$.
	
	\item We create the weighted average vector $\vec{z}' = \sum_{i=1}^{k} \frac{w_i}{w_{\mbox{\tiny \scshape{total}}}} \cdot \vec{z_i}$  in the embedding space, where $w_{\mbox{\tiny \scshape{total}}} = \sum_{j=1}^{k} w_j$.

	Several distance weighting schemes have been discussed in $k$-NN literature~\cite{aha1991instance}. In~our case we have employed two: 	(1) weight neighbors by 1 minus their distance ($w_i = 1 - d_i$)  and (2) weight neighbors by the inverse of their distance squared ($w_i = \frac{1}{d_i^2}$).
	
	\item The decoder is used to convert this average vector $\vec{z}'$ from the embedding space to the original label space as $y' = Dec(\vec{z}')$
	
{Various cutting and thresholding} 
 schemes can be used to binarize this vector and return the list of predicted~labels.
	\begin{itemize}
		\item Estimate the number of labels to return, $r$, from~the sizes of label sets of documents in $N(x)$, as~described in ~cite{cual}, and~return the $r$ predicted labels with the highest score.
		\item Apply a threshold on the activation of decoder output neurons to decide which labels have an excitation level high enough to be part of the final prediction.
	\end{itemize}
	
\end{enumerate}

\vspace{-6pt} {}
\unskip

\section{Results and~Discussion}~\label{sec:results}
This section conducts an exhaustive set of experiments on a large portion of the MEDLINE collection. In~these experiments we validate the effectiveness of our proposal of a multi-label $k$-NN text classifier assisted by a \LAE\ in a complex semantic indexing task. Different parameters and options were evaluated on the test dataset in order to determine the best setting for our system with the aim to answer the following research~questions:
\begin{itemize}
	\item What is the effect on classification performance of the choice of training document representations? Are there substantial differences between sparse term-based similarity and dense vector-based similarity?
	\item What are the best parameterizations for \LAEs\ (size of embedding representation layer, sizes of encoder and decoder layers, etc)?
	What are the effects of retrieving different number of neighbor documents on the classification performance and how affects the weighting scheme employed when creating the average embedded vector?
\end{itemize}

In this section we provide a description of our evaluation data and the performance metrics being employed and discuss the experimental results. The source code used to carry out the reported experiments is {available at } 
\url{https://github.com/fribadas/labelAE-MeSH}.

\begin{table}[H]
	\caption{{Evaluation dataset} 
 statistics and MeSH descriptor~distribution.}	
	\label{table:dataset} 
\newcolumntype{C}{>{\centering\arraybackslash}X}
\begin{tabularx}{\textwidth}{m{3.5cm}<{\raggedright}m{2.5cm}<{\centering}m{2.5cm}<{\centering}m{3cm}<{\centering}}
		\toprule
		\multicolumn{4}{c}{\textbf{{Collection statistics} 
}} \\
		\midrule
		\# citations                        &6,791,951 & & \\
		\# citations in dev dataset &10,000 & &  \\
		\# citations in test dataset        & 10,000 & &  \\
		\# MeSH descriptors                 & 29,483 & &  \\
		\midrule
		& min    & max     & avg \\
		descriptors per citation    & 1      & 19      & 12.90\\
		descriptor occurrences	    & 1      & 4,621,007 \textsuperscript{$\dagger$} & 2972.26\\  
	\midrule
		\multicolumn{4}{c}{\textbf{MeSH descriptor distribution}}\\
		\midrule
		\textbf{{occurrences} 
} & \multicolumn{3}{c}{\textbf{count}} \\   
		$\ge$ 1 M    & \multicolumn{3}{c}{7} \\
		$\ge$ 100 M  & \multicolumn{3}{c}{65}  \\
		$\ge$ 10 K   & \multicolumn{3}{c}{1314} \\
		$\ge$ 1 K    & \multicolumn{3}{c}{8752} \\
		$\ge$ 100   & \multicolumn{3}{c}{21,310}  \\                 
		\bottomrule
\end{tabularx}
\footnotesize{\textsuperscript{$\dagger$}  Descriptor D006801: \emph{Humans}}
\end{table}

\subsection{Dataset Details and Evaluation~Metrics}

Our experiments were conducted on a large textual multi-label dataset created as a subset of the 2021 edition of MEDLINE/PubMed {baseline files} 
~(\url{ftp://ftp.ncbi.nlm.nih.gov/pubmed/baseline}), which comprises over 6 million citations from 2010 onwards. 
For convenience, the~actual dataset was retrieved from the BioASQ challenge~\cite{tsatsaronis2015overview} {repository}~(\url{http://www.bioasq.org/}) rather than from the original sources.
BioASQ organizers retrieved the citations from the MEDLINE sources, extracting the relevant elements ({\scshape pmid}, {\scshape ArticleTitle}, {\scshape AbstractText}, {\scshape MeshHeadingsList}, {\scshape JournalName} and {\scshape Year}) and distributed then conveniently formatted as a JSON document. Table~\ref{table:dataset} summarizes the most relevant characteristics of the resulting~dataset.

\begin{table}[H]
	\caption{\label{table:autoencoders} {Configuration of label} 
 autoencoders in our~experiments.}

\begin{adjustwidth}{-\extralength}{0cm}
	\begin{center}
		\footnotesize
		\begin{tabularx}{\fulllength}{lCCCCCCr}
			\toprule	
			& \textbf{Input/Output}  &  \textbf{Encoder-1}   & \textbf{Encoder-2} & \textbf{Embedding} & \textbf{Decoder-1} & \textbf{Decoder-2}  & \textbf{\# Parameters} \\
			\midrule           
			\textsc{small}      & 29,483              &  1024     & 256    &   64         & 256      & 1024   &   60,975,467 \\
			\textsc{medium}    & 29,483              &  2048     & 512    &   128        & 512      & 2048   &   123,035,563 \\
			\textsc{large}      & 29,483              &  4096     & 1024  &   128        & 1024    & 4096   &   250,256,299 \\
			\bottomrule
		\end{tabularx}    
	\end{center}
\end{adjustwidth}

\end{table}

In our study we have employed two complementary sets of evaluation metrics that are commonly used in evaluating multi-label and XML~problems.
\begin{itemize}

	\item The evaluation of binary classifiers typically employs Precision (P), which measures how many predicted labels are correct, Recall (R), which counts how many correct labels the evaluated model is able to predict, and~F-score (F), which combines both metrics by calculating the harmonic mean of P and R.
In multi-class and multi-label problems these metrics are generalized by calculating their Macro-averaged and Micro-averaged variants.
A Macro-averaged measure computes a class-wide average of the corresponding measure while a Micro-averaged one computes the corresponding measure on all examples at once and, in~the general case, uses to have  the advantage of adequately handling the class imbalance.
In our evaluation we followed the BioASQ challenge proposal~\cite{tsatsaronis2015overview} that employs the Micro-averaged versions of Precision ($MiP$), Recall ($MiR$) and F-score ($MiF$) as main performance metrics, using $MiP$ as a ranking~criteria.

\item In XML, where the number of candidate labels is very large, metrics that focus on evaluating the effectiveness in predicting correct labels and generating an adequate ranking in the predicted label set are frequently used.
Precision at top $k$ ($P@k$) computes the fraction of correct predictions in the top $k$  predicted labels.
Normalized Discounted Cummulated Gain at top $k$ ($nDCG@k$)~\cite{jarvelin2002cumulated} is a measure of the ranking quality at the top $k$ predicted labels, which evaluates the usefulness of a predicted label according its position in the result list.
In our experimental results, we report the average $P@k$ and $nDCG@k$ on the testing set with $k = 5$ and $k=10$, in~order to provide a measure of prediction effectiveness.		
\end{itemize}

\subsection{Experimental~Results}

In the first place we have evaluated the performance of the different approaches described in Section~\ref{sec:representations} for document representations.
Secondly, another set of experiments has evaluated the performance of the $k$-NN method assisted with \LAEs\ described in Section~\ref{sec:knn_ae} using different AE~configurations.

\subsubsection{Dense vs. Sparse~Representations}

In order to evaluate the influence of the document representation being used on the categorization performance we have performed a battery of experiments comparing the use of the dense representations with contextual vectors (runs \textsc{dense}) and the use of different alternatives for extracting sparse representations.
In particular, we have evaluated the performance of single terms extracted by stemming (runs \textsc{Stems}) and lemmatization (runs \textsc{Lemmas}), the~combination of the different methods for extracting compound terms (runs \textsc{multi} where we combine \textsc{NERS}, \textsc{NPS} and \textsc{Keywords}) and the joint use of the terms extracted using all the methods described in Section~\ref{sec:representations} (runs \textsc{all}). The~effect of the number of neighbors considered in each case has also been evaluated, taking $k$ values in $\{5, 10, 20, 30, 50. 100\}$.

\begin{figure}[H]

\begin{adjustwidth}{-\extralength}{0cm}
\centering 
	\includegraphics[width=0.99\linewidth]{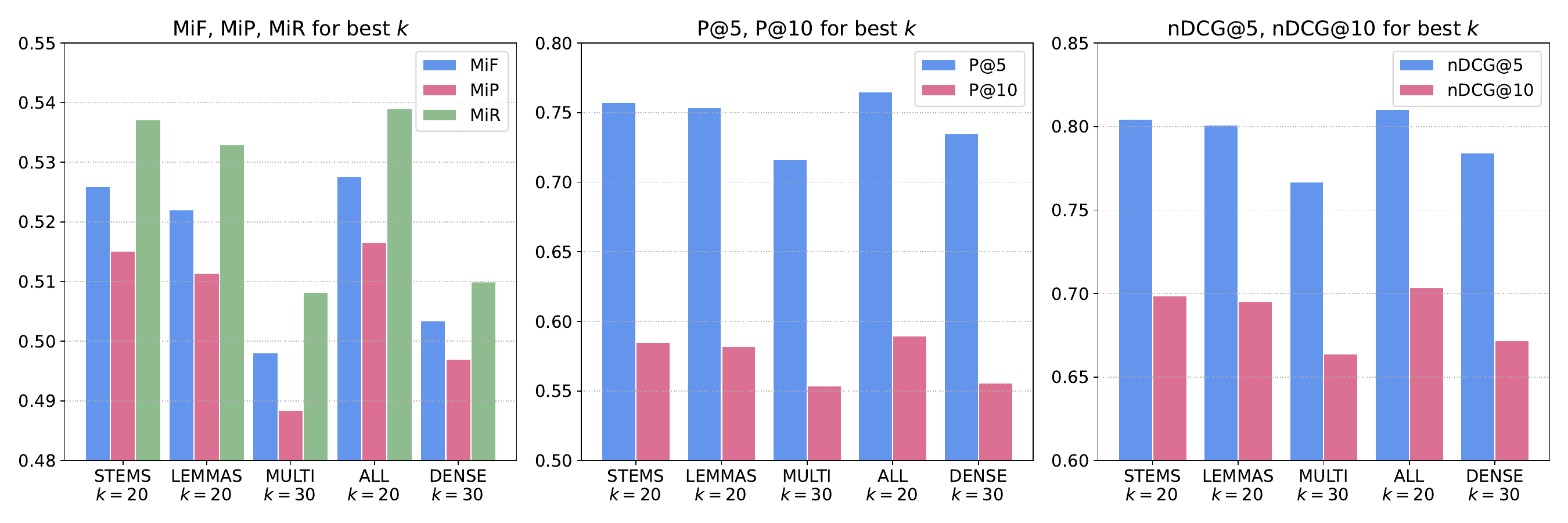}
\end{adjustwidth}
			\caption{\label{fig:sparse_dense} Summary of performance metrics 
				with sparse vs. dense representations for values of $k$ with best $MiF$ values.}

\end{figure}

As can be seen from the results shown in {Table} 
~\ref{tab:sparse_dense} and summarized in Figure~\ref{fig:sparse_dense}, for~these experiments the dense representation performs worse than most sparse representations in all performance metrics being considered and for all values of $k$.
The contribution of multi-word terms in the sparse representations is very limited. Although~the best results are obtained by combining all term extraction methods (runs \textsc{all}), it is observed that in all metrics the results obtained using single-word terms of type \textsc{Stems} and \textsc{Lemmas} dominate.
We hypothesize that when applying this kind of  $k$-NN method on a relatively large dataset ($>$6 M documents in our case) the contribution of more sophisticated representation methods is diluted.
In smaller datasets the use of very specific and precise multi-word terms can help to greatly improve the representation of a document when searching for similar~ones.

\vspace{-6pt} {}

\begin{table}[H]
	\begin{center}
	\caption{\label{tab:sparse_dense}Performance metrics with sparse vs. dense~representations.}
\begin{center}
\scriptsize
\setlength{\tabcolsep}{8pt}

\begin{tabularx}{\textwidth}{lrccccccc}
\toprule
                   
               & $\mathbf{k}$ & \textbf{MiF} & \textbf{MiP} & \textbf{MiR} & \textbf{P@5} & \textbf{P@10} & \textbf{nDCG@5} & \textbf{nDCG@10} \\ 
\midrule
\textsc{Stemms}
       &  5 & 0.4943 &  0.4855 &  0.5035 & 0.7299 & 0.5527 & 0.7706 & 0.6613 \\ 
      &  10 & 0.5191 &  0.5082 &  0.5305 & 0.7550 & 0.5804 & 0.7999 & 0.6925 \\ 
      &  20 & \bfseries {0.5259} 
 &  \bfseries 0.5151 &  \bfseries 0.5371 & 0.7574 & \bfseries 0.5849 & 0.8043 & \bfseries 0.6986 \\ 
      &  30 & 0.5240 &  0.5129 &  0.5355 & \bfseries 0.7579 & 0.5837 & \bfseries 0.8045 & 0.6974 \\ 
      &  50 & 0.5223 &  0.5111 &  0.5340 & 0.7527 & 0.5805 & 0.7991 & 0.6934 \\ 
      &  100 & 0.5147 &  0.5031 &  0.5269 & 0.7413 & 0.5731 & 0.7898 & 0.6857 \\ 
\midrule
\textsc{Lemmas}
       &  5 & 0.4920 &  0.4823 &  0.5020 & 0.7226 & 0.5496 & 0.7644 & 0.6574 \\ 
      &  10 & 0.5182 &  0.5078 &  0.5290 & 0.7485 & 0.5768 & 0.7931 & 0.6878 \\ 
      &  20 & \bfseries  0.5220 & \bfseries  0.5114 &  0.5330 & 0.7538 & 0.5819 & 0.8010 & 0.6951 \\ 
      &  30 & 0.5219 &  0.5107 & \bfseries  0.5336 & \bfseries 0.7544 & \bfseries 0.5822 & \bfseries 0.8018 & \bfseries 0.6959 \\ 
      &  50 & 0.5197 &  0.5087 &  0.5312 & 0.7488 & 0.5780 & 0.7960 & 0.6906 \\ 
      &  100 & 0.5143 &  0.5029 &  0.5262 & 0.7370 & 0.5704 & 0.7854 & 0.6821 \\ 
\midrule
\textsc{multi}
       &  5 & 0.4587 &  0.4492 &  0.4686 & 0.6909 & 0.5175 & 0.7329 & 0.6230 \\ 
      &  10 & 0.4875 &  0.4777 &  0.4977 & 0.7111 & 0.5441 & 0.7584 & 0.6529 \\ 
      &  20 & 0.4972 &  0.4875 &  0.5072 & \bfseries 0.7223 & \bfseries 0.5540 & \bfseries 0.7709 & \bfseries 0.6646 \\ 
      &  30 & \bfseries 0.4981 &  \bfseries 0.4884 &  \bfseries 0.5082 & 0.7165 & 0.5536 & 0.7670 & 0.6639 \\ 
      &  50 & 0.4945 &  0.4845 &  0.5049 & 0.7133 & 0.5500 & 0.7643 & 0.6605 \\ 
      &  100 & 0.4897 &  0.4796 &  0.5002 & 0.7026 & 0.5437 & 0.7531 & 0.6519 \\ 
\midrule
\textsc{all} &  5 & 0.4945 &  0.4856 &  0.5036 & 0.7276 & 0.5530 & 0.7681 & 0.6610 \\ 
      &  10 & 0.5207 &  0.5111 &  0.5307 & 0.7544 & 0.5795 & 0.8003 & 0.6930 \\ 
      &  20 & \bfseries 0.5276 &  \bfseries 0.5166 &  \bfseries 0.5390 & \bfseries 0.7649 & \bfseries 0.5894 & \bfseries 0.8103 & \bfseries 0.7035 \\ 
      &  30 & 0.5274 &  0.5163 &  0.5389 & 0.7611 & 0.5861 & 0.8079 & 0.7009 \\ 
      &  50 & 0.5237 &  0.5127 &  0.5352 & 0.7552 & 0.5821 & 0.8022 & 0.6958 \\ 
      &  100 & 0.5176 &  0.5064 &  0.5293 & 0.7453 & 0.5753 & 0.7933 & 0.6884 \\ 
\midrule
\textsc{dense} &  5 & 0.4779 &  0.4725 &  0.4834 & 0.7056 & 0.5299 & 0.7479 & 0.6380 \\ 
      &  10 & 0.4996 &  0.4936 &  0.5058 & 0.7348 & 0.5541 & 0.7800 & 0.6675 \\ 
      &  20 & 0.5030 &  \bfseries 0.4970 &  0.5093 & 0.7327 & \bfseries 0.5575 & 0.7826 & \bfseries 0.6728 \\ 
      &  30 & \bfseries 0.5034 &  \bfseries 0.4970 &  \bfseries 0.5100 & \bfseries 0.7350 & 0.5556 & \bfseries 0.7843 & 0.6718 \\ 
      &  50 & 0.5016 &  0.4950 &  0.5084 & 0.7291 & 0.5554 & 0.7789 & 0.6697 \\ 
      &  100 & 0.4918 &  0.4848 &  0.4991 & 0.7161 & 0.5451 & 0.7672 & 0.6586 \\ 
\bottomrule
\end{tabularx}
\end{center}

	\end{center}
\end{table}

In this context it is surprising that an apriori simpler approach such as the extraction of sparse representations and the use of the  Apache Lucene similarity performs better than the transformer-based contextual representations that currently dominate in the NLP research. An~in-depth review of this phenomenon is beyond the scope of this paper, it may be due to the lack of a prior fine-tuning phase with the employed MEDLINE dataset, a~poor suitability as a similarity metric of the Euclidean distance computed by the FAISS library or an inherent limitation of large pre-trained language models based on transformers as is discussed in~\cite{fi14010010}.

With respect to the number of neighbors to consider in the $k$-NN classification, the~best results are usually obtained with $k=20$ and $k=30$, which is in line with previous publications~\cite{trieschnigg2009mesh} in MeSH semantic~indexing.

\subsubsection{$k$-NN Prediction with Label~Autoencoders}

Regarding the experiments evaluating the performance of our proposal of a \LAE\  as a mechanism for improving $k$-NN classification, our objective has been to evaluate three aspects: (1) the performance of different \LAE\  topologies (2) the effect of the distance weighting scheme used to create the average vectors feeding the decoder and (3) the most appropriate threshold values to generate the list of predicted labels from the reconstruction of the label space provided by the~decoder.

Table~\ref{table:autoencoders} shows the characteristics of the \LAEs\ we have used in this series of experiments
We have employed a fixed neural network architecture, using two fully connected layers in both encoder and decoder and one fully connected layer as embedding layer.
We have trained and evaluated an encoder, named \textsc{small} \LAE, that uses a 64-dimensional embedding vector and an initial encoder and final decoder layer with 1024 neurons. We have also employed two AE architectures with a 128-dimensional embedding space with two encoder-decoder sizes, one with layers of 2048 and 256 neurons, called \textsc{medium} \LAE, and~another with encoder-decoder layers  of 4096 and 512 neurons, denoted as \textsc{large} \LAE.
We aimed to evaluate the effect of the size and the number of parameters in the learned \LAEs\ on their quality in the label encoding and reconstruction~tasks. 

\begin{figure}[H]
		\begin{adjustwidth}{-\extralength}{0cm}
	
				\includegraphics[width=1.0\linewidth]{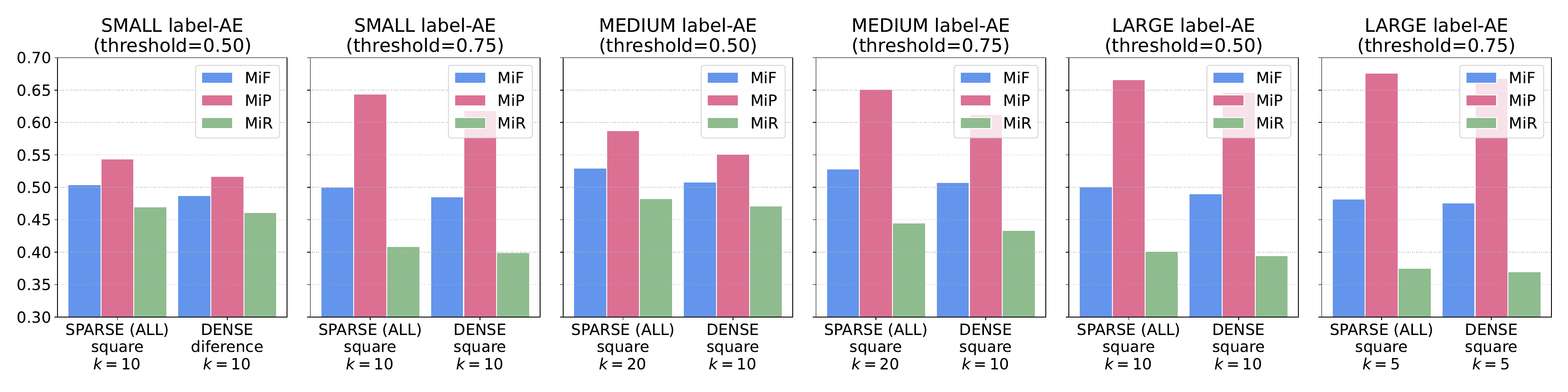}
					\end{adjustwidth}
				\caption{\label{fig:knn_ae} Summary of $MiF$, $MiP$, $MiR$ metrics 
					for values of $k$ and distance weighting with best $MiF$ values in each \LAE\ configuration (\textsc{small}, \textsc{medium}, \textsc{large}).}	
			\
	
	\end{figure}

The detailed results obtained with the \textsc{small} \LAE\ are shown in Table~\ref{tab:knn_ae_small}, those for the \textsc{medium} \LAE\ in Table~\ref{tab:knn_ae_medium} and those for the \textsc{large} \LAE\ in Table~\ref{tab:knn_ae_large}. 
Regarding the thresholds to be applied on the decoder output to create the list of predicted labels, two values have been evaluated, selecting those labels whose output activation exceed the value 0.5 in one case and the value 0.75 in the other. 
In this way we intended to evaluate the effect of considering more or less demanding selection criteria in conforming the predicted label list.
Finally, the~effect of the two distance weighting schemes introduced in Section~\ref{sec:knn_ae} to combine the embedded vectors has been evaluated in the different scenarios.
In both cases, weighting by 1 minus distance (\textsc{difference}) and weighting by the inverse of distance squared (\textsc{square}), the~\textsc{dense} representation and the \textsc{sparse} representation using all of the term extraction methods have been employed, using a number of neighbors  $k \in \{5, 10, 20, 30, 50, 100\}$.
Figure~\ref{fig:knn_ae} summarizes the $MiF$, $MiP$ and $MiR$ results for the best configurations of \LAE, threshold, distance weighting scheme and $k$.

\begin{table}[H]

		\caption{\label{tab:knn_ae_small} Performance with \textsc{small} \LAE.}
			\begin{adjustwidth}{-\extralength}{0cm}

\scriptsize
\setlength{\tabcolsep}{7.5pt}

\begin{tabularx}{\fulllength}{ccccccccccc}

     \toprule
                      
   \textbf{Threshold} & \textbf{Neighbors} & \textbf{Weighting}  & $\mathbf{k}$ & \textbf{MiF} & \textbf{MiP} & \textbf{MiR} & \textbf{P@5} & \textbf{P@10} & \textbf{nDCG@5} & \textbf{nDCG@10} \\ 
     \midrule
     $0.50$ & \textsc{sparse(all)} & \textsc{difference} &  5 & 0.4940 &  0.5338 &  0.4597 & 0.7042 & 0.5218 & 0.7492 & 0.6330      \\ 
           &       &      &   10 & \bfseries {0.4981} 
 &  \bfseries 0.5388 &  \bfseries 0.4632 & \bfseries 0.7153 & \bfseries 0.5256 & \bfseries 0.7620 & \bfseries 0.6408      \\ 
           &       &      &   20 & 0.4914 &  0.5299 &  0.4581 & 0.7094 & 0.5209 & 0.7566 & 0.6362      \\ 
           &       &      &   30 & 0.4886 &  0.5290 &  0.4540 & 0.7081 & 0.5173 & 0.7546 & 0.6322      \\ 
           &       &      &   50 & 0.4826 &  0.5250 &  0.4466 & 0.6992 & 0.5095 & 0.7475 & 0.6248      \\ 
           &       &      &   100 & 0.4719 &  0.5190 &  0.4326 & 0.6909 & 0.4956 & 0.7396 & 0.6120      \\ 
     \midrule
     $0.50$ & \textsc{sparse(all)} & \textsc{square} &  5 & 0.4950 &  0.5319 &  0.4629 & 0.7045 & 0.5221 & 0.7477 & 0.6318      \\ 
           &       &      &   10 & \bfseries 0.5038 &  \bfseries 0.5434 &  \bfseries 0.4696 & 0.7168 & \bfseries 0.5302 & 0.7625 & \bfseries 0.6439      \\ 
           &       &      &   20 & 0.4999 &  0.5399 &  0.4655 & 0.7189 & 0.5278 & \bfseries 0.7640 & 0.6426      \\ 
           &       &      &   30 & 0.4965 &  0.5367 &  0.4619 & \bfseries 0.7193 & 0.5256 & 0.7638 & 0.6405      \\ 
           &       &      &   50 & 0.4906 &  0.5329 &  0.4545 & 0.7114 & 0.5180 & 0.7581 & 0.6342      \\ 
           &       &      &   100 & 0.4810 &  0.5265 &  0.4427 & 0.7018 & 0.5048 & 0.7492 & 0.6218      \\ 
     \midrule
     $0.50$ & \textsc{dense} & \textsc{difference} &  5 & 0.4821 &  0.5140 &  0.4539 & 0.6929 & 0.5129 & 0.7390 & 0.6234      \\ 
           &       &      &   10 & \bfseries 0.4874 &  \bfseries 0.5168 &  \bfseries 0.4611 & 0.7019 & \bfseries 0.5207 & 0.7493 & \bfseries 0.6331      \\ 
           &       &      &   20 & 0.4857 &  0.5142 &  0.4601 & \bfseries 0.7026 & 0.5192 & \bfseries 0.7500 & 0.6321      \\ 
           &       &      &   30 & 0.4822 &  0.5101 &  0.4571 & 0.6987 & 0.5171 & 0.7467 & 0.6297      \\ 
           &       &      &   50 & 0.4755 &  0.5040 &  0.4501 & 0.6931 & 0.5104 & 0.7420 & 0.6236      \\ 
           &       &      &   100 & 0.4688 &  0.4969 &  0.4437 & 0.6850 & 0.5039 & 0.7349 & 0.6169      \\ 
     \midrule
     $0.50$ & \textsc{dense} & \textsc{square} &  5 & 0.4825 &  0.5142 &  0.4545 & 0.6943 & 0.5126 & 0.7402 & 0.6234      \\ 
           &       &      &   10 & \bfseries 0.4873 &  \bfseries 0.5168 &  \bfseries 0.4609 & 0.7024 & \bfseries 0.5202 & 0.7497 & \bfseries 0.6328      \\ 
           &       &      &   20 & 0.4850 &  0.5130 &  0.4599 & \bfseries 0.7045 & 0.5187 & \bfseries 0.7516 & 0.6321      \\ 
           &       &      &   30 & 0.4811 &  0.5089 &  0.4561 & 0.6954 & 0.5153 & 0.7440 & 0.6279      \\ 
           &       &      &   50 & 0.4734 &  0.5017 &  0.4481 & 0.6884 & 0.5069 & 0.7386 & 0.6204      \\ 
           &       &      &   100 & 0.4631 &  0.4903 &  0.4388 & 0.6782 & 0.4963 & 0.7291 & 0.6097      \\ 
   \midrule

     $0.75$ & \textsc{sparse(all)} & \textsc{difference} &  5 & 0.4893 &  0.6281 &  0.4007 & 0.6900 & 0.4836 & 0.7384 & 0.6024      \\ 
           &       &      &   10 & \bfseries 0.4956 &  \bfseries 0.6433 &  \bfseries 0.4030 & \bfseries 0.6985 & \bfseries 0.4869 & \bfseries 0.7495 & \bfseries 0.6100      \\ 
           &       &      &   20 & 0.4898 &  0.6411 &  0.3963 & 0.6928 & 0.4805 & 0.7441 & 0.6036      \\ 
           &       &      &   30 & 0.4857 &  0.6405 &  0.3911 & 0.6883 & 0.4749 & 0.7398 & 0.5982      \\ 
           &       &      &   50 & 0.4794 &  0.6397 &  0.3833 & 0.6787 & 0.4662 & 0.7320 & 0.5896      \\ 
           &       &      &   100 & 0.4677 &  0.6350 &  0.3702 & 0.6682 & 0.4508 & 0.7219 & 0.5749      \\ 
     \midrule
     $0.75$ & \textsc{sparse(all)} & \textsc{square} &  5 & 0.4915 &  0.6247 &  0.4051 & 0.6941 & 0.4874 & 0.7401 & 0.6051      \\ 
           &       &      &   10 & \bfseries 0.4998 &  0.6436 &  \bfseries 0.4085 & 0.7038 & \bfseries 0.4914 & 0.7527 & \bfseries 0.6131      \\ 
           &       &      &   20 & 0.4985 &  \bfseries 0.6496 &  0.4044 & \bfseries 0.7046 & 0.4889 & \bfseries 0.7532 & 0.6116      \\ 
           &       &      &   30 & 0.4941 &  0.6488 &  0.3989 & 0.7007 & 0.4840 & 0.7498 & 0.6071      \\ 
           &       &      &   50 & 0.4879 &  0.6460 &  0.3919 & 0.6922 & 0.4759 & 0.7437 & 0.6002      \\ 
           &       &      &   100 & 0.4773 &  0.6426 &  0.3796 & 0.6787 & 0.4608 & 0.7316 & 0.5855      \\ 
     \midrule
     $0.75$ & \textsc{dense} & \textsc{difference} &  5 & 0.4828 &  0.6109 &  0.3991 & 0.6828 & 0.4793 & 0.7316 & 0.5971      \\ 
           &       &      &   10 & 0.4842 &  0.6172 &  \bfseries 0.3983 & 0.6885 & 0.4813 & 0.7394 & 0.6020      \\ 
           &       &      &   20 & \bfseries 0.4850 &  \bfseries 0.6220 &  0.3974 & \bfseries 0.6900 & \bfseries 0.4814 & \bfseries 0.7407 & \bfseries 0.6025      \\ 
           &       &      &   30 & 0.4817 &  0.6196 &  0.3941 & 0.6851 & 0.4778 & 0.7367 & 0.5988      \\ 
           &       &      &   50 & 0.4753 &  0.6130 &  0.3881 & 0.6789 & 0.4712 & 0.7315 & 0.5926      \\ 
           &       &      &   100 & 0.4673 &  0.6083 &  0.3794 & 0.6692 & 0.4616 & 0.7232 & 0.5834      \\ 
 \midrule
     $0.75$ & \textsc{dense} & \textsc{square} &  5 & 0.4831 &  0.6106 &  0.3997 & 0.6836 & 0.4794 & 0.7323 & 0.5973      \\ 
           &       &      &   10 & \bfseries 0.4851 &  0.6182 &  \bfseries 0.3991 & 0.6893 & \bfseries 0.4820 & 0.7399 & \bfseries 0.6026      \\ 
           &       &      &   20 & 0.4842 &  \bfseries 0.6210 &  0.3967 & \bfseries 0.6911 & 0.4807 & \bfseries 0.7417 & 0.6022      \\ 
           &       &      &   30 & 0.4803 &  0.6176 &  0.3930 & 0.6818 & 0.4760 & 0.7340 & 0.5970      \\ 
           &       &      &   50 & 0.4718 &  0.6096 &  0.3849 & 0.6742 & 0.4677 & 0.7280 & 0.5894      \\ 
           &       &      &   100 & 0.4617 &  0.6031 &  0.3740 & 0.6611 & 0.4552 & 0.7163 & 0.5768      \\ 
     \bottomrule

\end{tabularx}

	\end{adjustwidth}
\end{table}

As can be seen in Table~\ref{tab:knn_ae_medium} the best results in all metrics are obtained with the \textsc{medium} \LAE, showing the 64-dimensional embedding space of the \textsc{small} \LAE\ as apparently incapable of adequately capturing relationships between MeSH descriptors and reconstructing them later.
The comparison between the performances of \textsc{medium} \LAE\ and \textsc{large} \LAE\ apparently confirms that 2048 dimensions in the input layer of the encoder are sufficient to provide an embedded representation capable, once reconstructed by the decoder, of~offering a performance in terms of $MiF$ similar to that of a basic $k$-NN method, improving its precision values at the cost of a slightly reduced recall.
Regarding the $P@k$ values and the measurement of ranking quality using $nDCG@k$, the~\LAE\  results are also able to equal those of the basic $k$-NN method. However, in~this case it is noteworthy that the \LAE\  method does not exceed the basic $k$-NN approach despite its good performance with respect to the $MiP$ metric.

\begin{table}[H]

		\caption{\label{tab:knn_ae_medium} Performance with \textsc{medium} \LAE.}
			\begin{adjustwidth}{-\extralength}{0cm}

\scriptsize
\setlength{\tabcolsep}{7pt}
\begin{tabularx}{\fulllength}{ccccccccccc}
     \toprule
                    
   \textbf{Threshold} & \textbf{Neighbors} & \textbf{Weighting}  & $\mathbf{k}$ & \textbf{MiF} & \textbf{MiP} & \textbf{MiR} & \textbf{P@5} & \textbf{P@10} & \textbf{nDCG@5} & \textbf{nDCG@10} \\ 
     \midrule
     $0.50$ & \textsc{sparse(all)} & \textsc{difference} &  5 & 0.5136 &  0.5532 &  0.4793 & 0.7265 & 0.5435 & 0.7717 & 0.6559      \\ 
           &       &      &   10 & \bfseries{ 0.5261} 
 &  0.5781 &  \bfseries 0.4827 & 0.7421 & \bfseries 0.5498 & 0.7896 & \bfseries 0.6681      \\ 
           &       &      &   20 & 0.5223 &  0.5845 &  0.4720 & \bfseries 0.7428 & 0.5444 & \bfseries 0.7914 & 0.6650      \\ 
           &       &      &   30 & 0.5208 &  \bfseries 0.5903 &  0.4660 & 0.7402 & 0.5396 & 0.7889 & 0.6606      \\ 
           &       &      &   50 & 0.5153 &  0.5915 &  0.4565 & 0.7347 & 0.5302 & 0.7841 & 0.6525      \\ 
           &       &      &   100 & 0.5056 &  0.5936 &  0.4403 & 0.7217 & 0.5157 & 0.7739 & 0.6396      \\ 
     \midrule
     $0.50$ & \textsc{sparse(all)} & \textsc{square} &  5 & 0.5114 &  0.5387 &  0.4867 & 0.7195 & 0.5435 & 0.7646 & 0.6539      \\ 
           &       &      &   10 & 0.5280 &  0.5716 &  \bfseries 0.4905 & 0.7420 & \bfseries 0.5556 & 0.7890 & 0.6718      \\ 
           &       &      &   20 & \bfseries 0.5298 &  0.5873 &  0.4826 & \bfseries 0.7492 & 0.5544 & \bfseries 0.7962 & \bfseries 0.6732      \\ 
           &       &      &   30 & 0.5274 &  \bfseries 0.5924 &  0.4752 & 0.7448 & 0.5477 & 0.7931 & 0.6677      \\ 
           &       &      &   50 & 0.5240 &  0.5967 &  0.4670 & 0.7432 & 0.5409 & 0.7919 & 0.6627      \\ 
           &       &      &   100 & 0.5155 &  0.5997 &  0.4521 & 0.7335 & 0.5282 & 0.7839 & 0.6513      \\ 
     \midrule
     $0.50$ & \textsc{dense} & \textsc{difference} &  5 & 0.4980 &  0.5279 &  0.4713 & 0.7109 & 0.5303 & 0.7566 & 0.6417      \\ 
           &       &      &   10 & \bfseries 0.5077 &  0.5514 &  \bfseries 0.4705 & 0.7233 & \bfseries 0.5356 & 0.7713 & \bfseries  0.6513      \\ 
           &       &      &   20 & 0.5074 &  \bfseries 0.5583 &  0.4650 & \bfseries 0.7234 & 0.5322 & \bfseries 0.7737 & 0.6507      \\ 
           &       &      &   30 & 0.5063 &  0.5613 &  0.4611 & 0.7194 & 0.5290 & 0.7708 & 0.6478      \\ 
           &       &      &   50 & 0.5002 &  0.5593 &  0.4524 & 0.7152 & 0.5216 & 0.7676 & 0.6418      \\ 
           &       &      &   100 & 0.4924 &  0.5573 &  0.4411 & 0.7061 & 0.5117 & 0.7592 & 0.6317      \\ 
     \midrule
     $0.50$ & \textsc{dense} & \textsc{square} 
                            &  5 & 0.4992 &  0.5287 &  \bfseries 0.4728 & 0.7113 & 0.5307 & 0.7567 & 0.6420      \\ 
           &       &      &   10 & \bfseries 0.5080 &  0.5512 &  0.4710 & \bfseries 0.7221 & \bfseries 0.5345 & 0.7702 & \bfseries 0.6502      \\ 
           &       &      &   20 & 0.5063 &  0.5569 &  0.4641 & 0.7211 & 0.5310 & \bfseries 0.7722 & 0.6497      \\ 
           &       &      &   30 & 0.5051 &  \bfseries 0.5607 &  0.4595 & 0.7179 & 0.5282 & 0.7700 & 0.6473      \\ 
           &       &      &   50 & 0.4977 &  0.5579 &  0.4492 & 0.7109 & 0.5178 & 0.7634 & 0.6377      \\ 
           &       &      &   100 & 0.4864 &  0.5532 &  0.4340 & 0.6983 & 0.5041 & 0.7527 & 0.6245      \\ 
  
     \midrule
     $0.75$ & \textsc{sparse(all)} & \textsc{difference} &  5 & 0.5152 &  0.6111 &  \bfseries 0.4453 & 0.7217 & 0.5264 & 0.7682 & 0.6426      \\ 
           &       &      &   10 & \bfseries 0.5251 &  0.6403 &  0.4450 & \bfseries 0.7348 & \bfseries 0.5288 & \bfseries 0.7841 & \bfseries 0.6515      \\ 
           &       &      &   20 & 0.5203 &  0\bfseries .6494 &  0.4340 & 0.7324 & 0.5198 & 0.7838 & 0.6454      \\ 
           &       &      &   30 & 0.5167 &  0.6560 &  0.4261 & 0.7285 & 0.5131 & 0.7801 & 0.6394      \\ 
           &       &      &   50 & 0.5090 &  0.6574 &  0.4152 & 0.7209 & 0.5016 & 0.7736 & 0.6291      \\ 
           &       &      &   100 & 0.4965 &  0.6587 &  0.3985 & 0.7076 & 0.4851 & 0.7630 & 0.6143      \\ 
     \midrule
     $0.75$ & \textsc{sparse(all)} & \textsc{square} 
                            &  5 & 0.5127 &  0.5943 &  0.4509 & 0.7163 & 0.5269 & 0.7622 & 0.6411      \\ 
           &       &      &   10 & 0.5265 &  0.6316 &  \bfseries 0.4514 & 0.7374 & \bfseries 0.5341 & 0.7858 & 0.6553      \\ 
           &       &      &   20 & \bfseries 0.5286 &  0.6512 &  0.4448 & \bfseries 0.7416 & 0.5320 & \bfseries 0.7907 & \bfseries 0.6557      \\ 
           &       &      &   30 & 0.5260 &  \bfseries 0.6594 &  0.4375 & 0.7365 & 0.5239 & 0.7868 & 0.6490      \\ 
           &       &      &   50 & 0.5189 &  0.6631 &  0.4261 & 0.7324 & 0.5128 & 0.7838 & 0.6404      \\ 
           &       &      &   100 & 0.5071 &  0.6651 &  0.4097 & 0.7191 & 0.4973 & 0.7731 & 0.6264      \\ 
     \midrule
     $0.75$ & \textsc{dense} & \textsc{difference} &  5 & 0.5018 &  0.5884 &  \bfseries 0.4374 & 0.7081 & 0.5139 & 0.7546 & 0.6292      \\ 
           &       &      &   10 & \bfseries 0.5072 &  0.6122 &  0.4330 & 0.7175 & \bfseries 0.5155 & 0.7670 & \bfseries 0.6356      \\ 
           &       &      &   20 & 0.5071 &  0.6245 &  0.4268 & \bfseries 0.7166 & 0.5103 & \bfseries 0.7685 & 0.6335      \\ 
           &       &      &   30 & 0.5026 &  \bfseries 0.6246 &  0.4204 & 0.7117 & 0.5047 & 0.7650 & 0.6288      \\ 
           &       &      &   50 & 0.4956 &  0.6257 &  0.4103 & 0.7052 & 0.4952 & 0.7601 & 0.6206      \\ 
           &       &      &   100 & 0.4875 &  0.6247 &  0.3998 & 0.6952 & 0.4849 & 0.7510 & 0.6104      \\ 
\bottomrule
\end{tabularx}
\end{adjustwidth}
   \end{table}

 \begin{table}[H]\ContinuedFloat
 \tablesize{\small}
\caption{{\em Cont.}}
			\begin{adjustwidth}{-\extralength}{0cm}

\scriptsize
\setlength{\tabcolsep}{7.5pt}

\begin{tabularx}{\fulllength}{ccccccccccc}

     \toprule
                      
   \textbf{Threshold} & \textbf{Neighbors} & \textbf{Weighting}  & $\mathbf{k}$ & \textbf{MiF} & \textbf{MiP} & \textbf{MiR} & \textbf{P@5} & \textbf{P@10} & \textbf{nDCG@5} & \textbf{nDCG@10} \\ 
     \midrule
     $0.75$ & \textsc{dense} & \textsc{square} &  5 & 0.5025 &  0.5881 &  \bfseries 0.4387 & 0.7082 & \bfseries 0.5145 & 0.7545 & 0.6297      \\ 
           &       &      &   10 & \bfseries 0.5075 &  0.6122 &  0.4334 & \bfseries 0.7164 & 0.5144 & 0.7660 & \bfseries 0.6346      \\ 
           &       &      &   20 & 0.5070 &  \bfseries 0.6250 &  0.4265 & 0.7147 & 0.5097 & \bfseries 0.7674 & 0.6330      \\ 
           &       &      &   30 & 0.5005 &  0.6233 &  0.4181 & 0.7104 & 0.5032 & 0.7643 & 0.6276      \\ 
           &       &      &   50 & 0.4918 &  0.6231 &  0.4062 & 0.7010 & 0.4910 & 0.7560 & 0.6164      \\ 
           &       &      &   100 & 0.4807 &  0.6198 &  0.3926 & 0.6869 & 0.4772 & 0.7440 & 0.6030      \\ 
     \bottomrule
  \end{tabularx}

	\end{adjustwidth}
\end{table}

With respect to the thresholds, both values have similar performance without great differences, being slightly better to prefer the stricter output criterion provided by the value 0.75.
Performance with sparse representations is still better than with dense context vectors, and~there is a slight tendency to get better results using fewer neighbors than the basic $k$-NN method.
Finally, the~results using the inverse of distance squared as the distance weighting scheme are superior in all scenarios, because~it boosts the contribution of the most similar examples when constructing the average embedded~vector.

\begin{table}[H]

		\caption{\label{tab:knn_ae_large} Performance with \textsc{large} \LAE.}
			\begin{adjustwidth}{-\extralength}{0cm}

\scriptsize
\setlength{\tabcolsep}{7.5pt}
\begin{tabularx}{\fulllength}{ccccccccccc}
     \toprule
                     
   \textbf{Threshold} & \textbf{Neighbors} & \textbf{Weighting}  & $\mathbf{k}$ & \textbf{MiF} & \textbf{MiP} & \textbf{MiR} & \textbf{P@5} & \textbf{P@10} & \textbf{nDCG@5} & \textbf{nDCG@10} \\ 
     \midrule
     $0.50$ & \textsc{sparse(all)} & \textsc{difference}
                            &  5 & \bfseries {0.4962 } 
&  0.6457 &  \bfseries 0.4029 & 0.6934 & \bfseries 0.4866 & 0.7439 & 0.6077      \\ 
           &       &      &   10 & 0.4959 &  0.6733 &  0.3925 & \bfseries 0.7006 & 0.4796 & \bfseries 0.7558 & \bfseries 0.6082      \\ 
           &       &      &   20 & 0.4899 &  0.6857 &  0.3812 & 0.6971 & 0.4702 & 0.7534 & 0.6011      \\ 
           &       &      &   30 & 0.4843 &  0.6901 &  0.3731 & 0.6890 & 0.4621 & 0.7469 & 0.5935      \\ 
           &       &      &   50 & 0.4747 &  0.6932 &  0.3610 & 0.6778 & 0.4488 & 0.7370 & 0.5811      \\ 
           &       &      &   100 & 0.4601 &  \bfseries 0.6981 &  0.3431 & 0.6609 & 0.4297 & 0.7226 & 0.5631      \\ 
     \midrule
     $0.50$ & \textsc{sparse(all)} & \textsc{square} 
                            &  5 & 0.4973 &  0.6285 &  \bfseries 0.4114 & 0.6924 & \bfseries 0.4952 & 0.7427 & \bfseries 0.6132      \\ 
           &       &      &   10 & \bfseries 0.5006 &  0.6662 &  0.4010 & 0.7064 & 0.4860 & 0.7584 & 0.6123      \\ 
           &       &      &   20 & 0.4972 &  0.6864 &  0.3897 & \bfseries 0.7078 & 0.4792 & \bfseries 0.7613 & 0.6090      \\ 
           &       &      &   30 & 0.4921 &  0.6917 &  0.3819 & 0.7030 & 0.4715 & 0.7580 & 0.6030      \\ 
           &       &      &   50 & 0.4835 &  0.6960 &  0.3704 & 0.6896 & 0.4594 & 0.7473 & 0.5917      \\ 
           &       &      &   100 & 0.4700 &  \bfseries 0.7017 &  0.3534 & 0.6727 & 0.4406 & 0.7333 & 0.5742      \\ 
     \midrule
     $0.50$ & \textsc{dense} & \textsc{difference}
                            &  5 & 0.4874 &  0.6179 &  \bfseries 0.4024 & 0.6889 & \bfseries 0.4859 & 0.7414 &\bfseries  0.6065      \\ 
           &       &      &   10 & \bfseries 0.4897 &  0.6473 &  0.3939 & \bfseries 0.6943 & 0.4801 & \bfseries 0.7496 & 0.6063      \\ 
           &       &      &   20 & 0.4868 &  0.6609 &  0.3853 & 0.6911 & 0.4746 & 0.7479 & 0.6024      \\ 
           &       &      &   30 & 0.4822 &  0.6646 &  0.3783 & 0.6879 & 0.4666 & 0.7453 & 0.5958      \\ 
           &       &      &   50 & 0.4754 &  \bfseries 0.6667 &  0.3694 & 0.6797 & 0.4572 & 0.7386 & 0.5873      \\ 
           &       &      &   100 & 0.4659 &  0.6661 &  0.3582 & 0.6696 & 0.4450 & 0.7305 & 0.5763      \\ 
     \midrule
     $0.50$ & \textsc{dense} & \textsc{square}
                            &  5 & 0.4884 &  0.6181 &  \bfseries 0.4036 & 0.6894 & \bfseries 0.4868 & 0.7414 & \bfseries 0.6069      \\ 
           &       &      &   10 & \bfseries 0.4899 &  0.6466 &  0.3944 & \bfseries 0.6946 & 0.4803 & \bfseries 0.7498 & 0.6065      \\ 
           &       &      &   20 & 0.4862 &  0.6606 &  0.3846 & 0.6894 & 0.4733 & 0.7462 & 0.6008      \\ 
           &       &      &   30 & 0.4814 &  0.6642 &  0.3776 & 0.6867 & 0.4651 & 0.7439 & 0.5941      \\ 
           &       &      &   50 & 0.4723 &  \bfseries 0.6656 &  0.3660 & 0.6749 & 0.4535 & 0.7352 & 0.5839      \\ 
           &       &      &   100 & 0.4600 &  0.6629 &  0.3521 & 0.6627 & 0.4383 & 0.7241 & 0.5694      \\ 

     \midrule
     $0.75$ & \textsc{sparse(all)} & \textsc{difference} 
                            &  5 & \bfseries 0.4795 &  0.6912 &  \bfseries 0.3671 & \bfseries 0.6779 & \bfseries 0.4541 & 0.7322 & \bfseries 0.5811      \\ 
           &       &      &   10 & 0.4755 &  0.7212 &  0.3546 & 0.6772 & 0.4424 & \bfseries 0.7377 & 0.5769      \\ 
           &       &      &   20 & 0.4677 &  0.7358 &  0.3428 & 0.6655 & 0.4296 & 0.7280 & 0.5651      \\ 
           &       &      &   30 & 0.4616 &  0.7450 &  0.3344 & 0.6584 & 0.4209 & 0.7228 & 0.5577      \\ 
           &       &      &   50 & 0.4508 &  0.7481 &  0.3226 & 0.6446 & 0.4064 & 0.7111 & 0.5441      \\ 
           &       &      &   100 & 0.4339 &  \bfseries 0.7511 &  0.3051 & 0.6262 & 0.3868 & 0.6945 & 0.5245      \\

\bottomrule
\end{tabularx}
\end{adjustwidth}
   \end{table}

 \begin{table}[H]\ContinuedFloat
 \tablesize{\small}
\caption{{\em Cont.}}
			\begin{adjustwidth}{-\extralength}{0cm}

\scriptsize
\setlength{\tabcolsep}{7.5pt}

\begin{tabularx}{\fulllength}{ccccccccccc}

     \toprule
                      
   \textbf{Threshold} & \textbf{Neighbors} & \textbf{Weighting}  & $\mathbf{k}$ & \textbf{MiF} & \textbf{MiP} & \textbf{MiR} & \textbf{P@5} & \textbf{P@10} & \textbf{nDCG@5} & \textbf{nDCG@10} \\ 
     \midrule

     $0.75$ & \textsc{sparse(all)} & \textsc{square}
                            &  5 & \bfseries 0.4822 &  0.6758 &  \bfseries 0.3749 & 0.6792 & \bfseries 0.4611 & 0.7328 & \bfseries 0.5863      \\ 
           &       &      &   10 & 0.4821 &  0.7143 &  0.3638 & \bfseries 0.6854 & 0.4510 & \bfseries 0.7427 & 0.5836      \\ 
           &       &      &   20 & 0.4771 &  0.7378 &  0.3526 & 0.6819 & 0.4418 & 0.7412 & 0.5770      \\ 
           &       &      &   30 & 0.4714 &  0.7468 &  0.3444 & 0.6735 & 0.4322 & 0.7353 & 0.5689      \\ 
           &       &      &   50 & 0.4611 &  0.7510 &  0.3326 & 0.6599 & 0.4190 & 0.7239 & 0.5564      \\ 
           &       &      &   100 & 0.4432 &  \bfseries 0.7532 &  0.3140 & 0.6373 & 0.3974 & 0.7053 & 0.5360      \\ 
     \midrule
     $0.75$ & \textsc{dense} & \textsc{difference}
                            &  5 & \bfseries 0.4746 &  0.6679 &  \bfseries 0.3681 & 0.6741 & \bfseries 0.4541 & 0.7304 & \bfseries 0.5811      \\ 
           &       &      &   10 & 0.4729 &  0.6971 &  0.3578 & \bfseries 0.6771 & 0.4461 & 0.\bfseries 7369 & 0.5790      \\ 
           &       &      &   20 & 0.4675 &  0.7157 &  0.3471 & 0.6691 & 0.4360 & 0.7314 & 0.5706      \\ 
           &       &      &   30 & 0.4623 &  0.7186 &  0.3408 & 0.6643 & 0.4289 & 0.7277 & 0.5645      \\ 
           &       &      &   50 & 0.4545 &  0.7199 &  0.3321 & 0.6533 & 0.4185 & 0.7188 & 0.5550      \\ 
           &       &      &   100 & 0.4448 &  \bfseries 0.7226 &  0.3212 & 0.6410 & 0.4052 & 0.7085 & 0.5425      \\ 
     \midrule
     $0.75$ & \textsc{dense} & \textsc{square}
                            &  5 & \bfseries 0.4761 &  0.6681 &  \bfseries 0.3698 & 0.6754 & \bfseries 0.4560 & 0.7309 & \bfseries 0.5822      \\ 
           &       &      &   10 & 0.4734 &  0.6983 &  0.3581 & \bfseries 0.6760 & 0.4461 & \bfseries 0.7361 & 0.5788      \\ 
           &       &      &   20 & 0.4668 &  0.7149 &  0.3466 & 0.6672 & 0.4346 & 0.7295 & 0.5690      \\ 
           &       &      &   30 & 0.4621 &  0.7197 &  0.3403 & 0.6626 & 0.4280 & 0.7256 & 0.5630      \\ 
           &       &      &   50 & 0.4516 &  0.7182 &  0.3294 & 0.6498 & 0.4152 & 0.7162 & 0.5519      \\ 
           &       &      &   100 & 0.4381 &  \bfseries 0.7215 &  0.3146 & 0.6328 & 0.3975 & 0.7010 & 0.5344      \\ 
     \bottomrule
  \end{tabularx}

	\end{adjustwidth}
\end{table}

When comparing the \textsc{medium} \LAE\ best results from Table~\ref{tab:knn_ae_medium} with the basic $k$-NN best results from Table~\ref{tab:sparse_dense} we can see that they show very similar $MiF$ values, which in the case of the \textsc{medium} \LAE\ model is obtained with relatively high values of $MiP$ at the expense of lower values in $MiR$, whereas the basic $k$-NN method offers values more uniform in both metrics.
After a detailed analysis of the predictions made by both models we have found that the number of labels predicted by the \textsc{medium} \LAE\ model is substantially smaller. 
In our study we have obtained that the average number of labels predicted for each document by the simple $k$-NN method is 13.34 for the sparse representation and 13.13 for the dense representation. 
For the \textsc{medium} \LAE\ model with a threshold of 0.50 its average length is 10.44 labels with the sparse representation and 10.61 with the dense representation, whereas with a threshold of 0.75 we have, respectively, 8.65 and 8.69 labels. 
This behavior makes the basic $k$-NN method start with an initial advantage in providing better values for $MiR$. 

We hypothesize that the proposed \LAE\ assisted $k$-NN method is capable of (1) providing faithful embedded vectors via its encoder and (2) acceptably reconstructing the output labels from the averaged embedded vectors using its decoder, hence offering high values in $MiP$, but~it leaves behind the less frequent labels, which have few training examples to assert their presence in the encoder and decoder weights.
On the other hand, the~results seem to indicate that the basic $k$-NN method is able to satisfactorily circumvent the treatment of infrequent labels, at~least in a large datasets such as the one we are dealing with. This is probably due to the very nature of the $k$-NN method. These infrequent labels appear in documents with very specific contents, which leads to a very particular set of neighbors that target the $k$-NN classifier to these rare~labels.

In order to try to combine the best aspects of both approaches, which are the high $MiP$ of our proposed $k$-NN with \LAE\  and the best recall capabilities of the classical $k$-NN method, we have carried out a battery of additional tests.
For this purpose, we have taken as a starting point the labels predicted  by the \LAE\ method and combined them with the predictions provided by the basic $k$-NN method.
To build the final set of output labels, to~the labels predicted by the \LAE\ based method we add labels taken from the basic $k$-NN prediction until the number of output labels predicted by the basic $k$-NN is~reached.

Table~\ref{tab:knn_mix_medium} shows the results obtained by combining according to the described scheme the predictions of the basic $k$-NN model with the predictions provided by the \textsc{medium} \LAE. In~the case of the $MiF$, $MiP$ and $MiR$ metrics all of them are substantially improved with respect to the values obtained with these methods separately. In~contrast, the~values of $P@k$ and $nDCG@k$ are~penalized.


	Although these results improve those provided by the basic $k$-NN method and those of the $k$-NN method assisted with our \LAE, they are far from those offered by the best state-of-the-art semantic indexing systems for MeSH. If~we take as a reference the latest editions of the BioASQ challenge~\cite{tsatsaronis2015overview, bioasq2021}, which proposes an evaluation scenario very similar to the one presented in this work, we see that the best systems are capable of reaching $MiF$ scores somewhat higher than 70\%, while the Default MTI (Medical Text Indexer) reached values between 53\% in the first edition of the challenge and values in the range  62--64\% in the last two editions. The~baseline used in the first editions of this challenge, which performed a simple string match of the label text, reached values around 26\%.

\begin{table}[H]

		\caption{\label{tab:knn_mix_medium}Performance mixing results from basic $k$-NN with \textsc{medium} \LAE.}
			\begin{adjustwidth}{-\extralength}{0cm}

\scriptsize
\setlength{\tabcolsep}{7.5pt}

\begin{tabularx}{\fulllength}{ccccccccccc}

     \toprule
   \textbf{Threshold} & \textbf{Neighbors} & \textbf{Weighting}  & $\mathbf{k}$ & \textbf{MiF} & \textbf{MiP} & \textbf{MiR} & \textbf{P@5} & \textbf{P@10} & \textbf{nDCG@5} & \textbf{nDCG@10} \\ 
     \midrule
     $0.50$ & \textsc{sparse(all)} & \textsc{difference} &  5 & 0.5109 &  0.5018 &  0.5204 & 0.5159 & 0.5076 & 0.5152 & 0.5341      \\ 
           &       &      &   10 & 0.5309 &  0.5208 &  0.5414 & 0.5384 & 0.5301 & 0.5398 & 0.5598      \\ 
           &       &      &   20 & \bfseries {0.5345} 
 &  \bfseries 0.5236 &  \bfseries 0.5460 & 0.5530 & \bfseries 0.5383 & 0.5534 & 0.5694      \\ 
           &       &      &   30 & 0.5330 &  0.5218 &  0.5448 & 0.5554 & 0.5382 & 0.5564 & \bfseries 0.5706      \\ 
           &       &      &   50 & 0.5304 &  0.5191 &  0.5422 & 0.5571 & 0.5371 & \bfseries 0.5586 & 0.5705      \\ 
           &       &      &   100 & 0.5243 &  0.5130 &  0.5361 & \bfseries 0.5590 & 0.5346 & 0.5605 & 0.5694      \\ 
     \midrule
     $0.50$ & \textsc{sparse(all)} & \textsc{square} &  5 & 0.5063 &  0.4973 &  0.5155 & 0.4993 & 0.4968 & 0.4987 & 0.5205      \\ 
           &       &      &   10 & 0.5294 &  0.5197 &  0.5395 & 0.5306 & 0.5277 & 0.5303 & 0.5536      \\ 
           &       &      &   20 & 0.5385 &  0.5277 &  0.5498 & 0.5484 & 0.5404 & 0.5496 & 0.5700      \\ 
           &       &      &   30 & \bfseries 0.5408 &  \bfseries 0.5299 &  \bfseries 0.5522 & 0.5562 & \bfseries 0.5432 & 0.5554 & 0.5732      \\ 
           &       &      &   50 & 0.5399 &  0.5286 &  0.5517 & \bfseries 0.5578 & 0.5429 & \bfseries 0.5583 & \bfseries 0.5743      \\ 
           &       &      &   100 & 0.5355 &  0.5239 &  0.5476 & 0.5555 & 0.5390 & 0.5537 & 0.5688      \\ 
     \midrule
     $0.50$ & \textsc{dense} & \textsc{difference} &  5 & 0.4942 &  0.4885 &  0.5002 & 0.4825 & 0.4864 & 0.4803 & 0.5065      \\ 
           &       &      &   10 & 0.5098 &  0.5038 &  0.5159 & 0.5050 & 0.5037 & 0.5033 & 0.5275      \\ 
           &       &      &   20 & \bfseries 0.5152 &  \bfseries 0.5089 &  \bfseries 0.5217 & 0.5101 & 0.5116 & 0.5071 & 0.5339      \\ 
           &       &      &   30 & 0.5148 &  0.5085 &  0.5212 & \bfseries 0.5130 & \bfseries 0.5134 & \bfseries 0.5093 & \bfseries 0.5359      \\ 
           &       &      &   50 & 0.5122 &  0.5059 &  0.5187 & 0.5080 & 0.5095 & 0.5070 & 0.5333      \\ 
           &       &      &   100 & 0.5067 &  0.4998 &  0.5139 & 0.5078 & 0.5065 & 0.5042 & 0.5290      \\ 
     \midrule
     $0.50$ & \textsc{dense} & \textsc{square} &  5 & 0.4944 &  0.4886 &  0.5003 & 0.4828 & 0.4854 & 0.4804 & 0.5056      \\ 
           &       &      &   10 & 0.5095 &  0.5036 &  0.5156 & 0.5043 & 0.5033 & 0.5033 & 0.5273      \\ 
           &       &      &   20 & \bfseries 0.5146 &  \bfseries 0.5083 &  \bfseries 0.5211 & 0.5086 & 0.5112 & 0.5053 & 0.5330      \\ 
           &       &      &   30 & 0.5136 &  0.5072 &  0.5200 & \bfseries 0.5096 & \bfseries 0.5113 & \bfseries 0.5067 & \bfseries 0.5339      \\ 
           &       &      &   50 & 0.5088 &  0.5022 &  0.5155 & 0.5066 & 0.5068 & 0.5041 & 0.5296      \\ 
           &       &      &   100 & 0.4993 &  0.4921 &  0.5066 & 0.5021 & 0.5007 & 0.4979 & 0.5227      \\ 

     \midrule
     $0.75$ & \textsc{sparse(all)} & \textsc{difference} &  5 & 0.5135 &  0.5043 &  0.5230 & 0.5706 & 0.5354 & 0.5694 & 0.5714      \\ 
           &       &      &   10 & 0.5346 &  0.5244 &  0.5452 & 0.5984 & 0.5612 & 0.5997 & 0.6010      \\ 
           &       &      &   20 & \bfseries 0.5412 &  0.5301 &  \bfseries 0.5528 & 0.6138 & 0.5692 & 0.6163 & 0.6124      \\ 
           &       &      &   30 & 0.5409 & \bfseries  0.5295 &  \bfseries 0.5528 & \bfseries  0.6199 & \bfseries 0.5712 & \bfseries 0.6233 & \bfseries 0.6166      \\ 
           &       &      &   50 & 0.5377 &  0.5262 &  0.5496 & 0.6199 & 0.5697 & 0.6227 & 0.6147      \\ 
           &       &      &   100 & 0.5308 &  0.5193 &  0.5427 & 0.6153 & 0.5625 & 0.6206 & 0.6097      \\ 
     \midrule
     $0.75$ & \textsc{sparse(all)} & \textsc{square} &  5 & 0.5055 &  0.4966 &  0.5148 & 0.5500 & 0.5235 & 0.5478 & 0.5546      \\ 
           &       &      &   10 & 0.5321 &  0.5223 &  0.5423 & 0.5873 & 0.5549 & 0.5883 & 0.5923      \\ 
           &       &      &   20 & 0.5446 &  0.5337 &  0.5560 & 0.6108 & 0.5729 & 0.6126 & 0.6136      \\ 
           &       &      &   30 & 0.5483 &  0.5372 &  0.5599 & 0.6172 & \bfseries 0.5768 & 0.6208 & \bfseries 0.6198      \\ 
           &       &      &   40 & \bfseries 0.5492 &  \bfseries 0.5378 &  0.5612 & \bfseries 0.6193 & 0.5755 & \bfseries 0.6217 & 0.6188      \\ 
           &       &      &   50 & 0.5482 &  0.5367 &  \bfseries 0.5601 & 0.6185 & 0.5743 & 0.6208 & 0.6173      \\ 
           &       &      &   100 & 0.5430 &  0.5313 &  0.5552 & 0.6080 & 0.5654 & 0.6086 & 0.6064      \\ 

\bottomrule
\end{tabularx}
\end{adjustwidth}
   \end{table}

 \begin{table}[H]\ContinuedFloat
 \tablesize{\small}
\caption{{\em Cont.}}
			\begin{adjustwidth}{-\extralength}{0cm}

\scriptsize
\setlength{\tabcolsep}{7.5pt}

\begin{tabularx}{\fulllength}{ccccccccccc}

     \toprule
                      
   \textbf{Threshold} & \textbf{Neighbors} & \textbf{Weighting}  & $\mathbf{k}$ & \textbf{MiF} & \textbf{MiP} & \textbf{MiR} & \textbf{P@5} & \textbf{P@10} & \textbf{nDCG@5} & \textbf{nDCG@10} \\ 
     \midrule

     $0.75$ & \textsc{dense} & \textsc{difference} &  5 & 0.4952 &  0.4894 &  0.5011 & 0.5384 & 0.5143 & 0.5346 & 0.5433      \\ 
           &       &      &   10 & 0.5133 &  0.5073 &  0.5194 & 0.5663 & 0.5320 & 0.5628 & 0.5660      \\ 
           &       &      &   20 & \bfseries 0.5221 &  \bfseries 0.5157 &  \bfseries 0.5287 & 0.5756 & \bfseries 0.5446 & 0.5717 & 0.5778      \\ 
           &       &      &   30 & 0.5201 &  0.5138 &  0.5266 & 0.5744 & 0.5443 & 0.5717 & \bfseries 0.5781      \\ 
           &       &      &   50 & 0.5181 &  0.5117 &  0.5246 & \bfseries 0.5768 & 0.5433 & \bfseries 0.5727 & 0.5773      \\ 
           &       &      &   100 & 0.5146 &  0.5075 &  0.5219 & 0.5751 & 0.5378 & 0.5695 & 0.5718      \\ 
     \midrule
     $0.75$ & \textsc{dense} & \textsc{square} &  5 & 0.4951 &  0.4894 &  0.5011 & 0.5387 & 0.5134 & 0.5335 & 0.5416      \\ 
           &       &      &   10 & 0.5142 &  0.5082 &  0.5203 & 0.5654 & 0.5320 & 0.5616 & 0.5654      \\ 
           &       &      &   20 & \bfseries 0.5211 &  \bfseries 0.5147 &  \bfseries 0.5277 & \bfseries 0.5757 & \bfseries 0.5443 & \bfseries 0.5722 & \bfseries 0.5780      \\ 
           &       &      &   30 & 0.5200 &  0.5136 &  0.5266 & 0.5722 & 0.5425 & 0.5690 & 0.5759      \\ 
           &       &      &   50 & 0.5155 &  0.5088 &  0.5223 & 0.5730 & 0.5378 & 0.5691 & 0.5725      \\ 
           &       &      &   100 & 0.5085 &  0.5013 &  0.5160 & 0.5686 & 0.5316 & 0.5633 & 0.5654      \\ 
     \bottomrule
  \end{tabularx}

	\end{adjustwidth}
\end{table}
\unskip

\section{Conclusions and Future~Work}~\label{sec:conclusions}
In this paper we propose a novel multi-label text categorization method able to deal with a very large and structured label space, that it is suitable to be applied in semantic indexing tasks using controlled vocabularies, such it is the case of the Medical Subject Headings (MeSH) thesaurus.
The proposed method trains a large \LAE\  capable of simultaneously learning an encoder function that transforms the original label space into a reduced-dimensional space, along with a decoder function that transforms vectors from that space back into the original label space.
The proposal adapts classical $k$-NN categorization to work in the semantic latent space learned by this \LAE.

We have proposed and  evaluated several document representation approaches, using both sparse textual features and dense contextual representations.
We have evaluated their contribution in finding neighboring documents employed in the $k$-NN~classification.

An exhaustive study on a large portion of MEDLINE collection has been carried out to evaluate different strategies in the definition and training of \LAEs\ for the MeSH thesaurus and to verify the suitability of the proposed classification method.
The results obtained confirm the ability of the learned \LAEs\ to capture the latent semantics of MeSH thesaurus descriptors and leverage that representation space in the $k$-NN~classification.

As a future work, a~direct application of the method described in this paper is to test the usefulness of the \LAEs\ learned for MeSH on related thesauri in other languages. An~example of such a thesaurus is the DeCS (\emph{Descriptores en Ciencias de la Salud}, Health Sciences Descriptors)  controlled {vocabulary} 
~(\url{http://decs.bvsalud.org/}), which is a trilingual~({In Portuguese, Spanish and English}) version of MeSH, retaining its structure and adding a collection of specific descriptors.
We hypothesize that it is possible to leverage the semantic information about MeSH condensed in the learned encoders and decoders to advantage of it in multilingual biomedical~environments.




\authorcontributions{%
Conceptualization, F.J.R.-P.; 
software, F.J.R.-P., V.M.D.B. and S.C.; 
validation, F.J.R.-P. and V.M.D.B.; 
investigation, F.J.R.-P.; 
resources, F.J.R.-P. and V.M.D.B.; 
data curation, F.J.R.-P., V.M.D.B. and S.C.; 
writing---original draft preparation, F.J.R.-P., V.M.D.B. and S.C.; 
writing---review and editing, F.J.R.-P.
supervision, F.J.R.-P. 
All authors have read and agreed to the published version of the~manuscript.
}

\funding{This research was partially funded by the Spanish Ministry of
	Science and Innovation through the project~PID2020-113230RB-C22.}



\institutionalreview{Not applicable.}

\informedconsent{Not applicable.}

\dataavailability{Data was obtained from the repository of the BioASQ challenge~\cite{tsatsaronis2015overview} and are available under {registration at} 
 \url{http://participants-area.bioasq.org/datasets/	}.}

\conflictsofinterest{The authors declare no conflict of~interest.}

\begin{adjustwidth}{-\extralength}{0cm}

\reftitle{References}

\end{adjustwidth}

\end{document}